\documentclass[journal,twoside,web]{IEEEcolor}
\usepackage{generic}
\usepackage{cite}
\usepackage{amsmath,amssymb,amsfonts}
\usepackage[ruled,vlined,noend]{algorithm2e}
\usepackage{graphicx}
\usepackage{textcomp}

\usepackage[]{units}
\usepackage{bm}
\usepackage{float}
\usepackage{amssymb}
\usepackage{url}


\usepackage{caption}
\usepackage{subcaption}
\usepackage{algpseudocode}
\usepackage[]{units}
\usepackage{paralist}
\usepackage{mathtools}

\usepackage{siunitx}
\usepackage{multirow}
\usepackage{epstopdf}
\graphicspath{ {Pictures/} }
\usepackage{epstopdf}
\usepackage{epsfig}
\usepackage{float}
\usepackage{hyperref}
\usepackage{amsmath} 
\usepackage{amssymb}  
\usepackage{mathtools}
\graphicspath{ {Pictures/} }
\DeclareGraphicsExtensions{.pdf,.eps}

\usepackage[acronym,nomain,nonumberlist]{glossaries}
\newacronym{uav}{UAV}{Unmanned Aerial Vehicle}
\newacronym{nmpc}{NMPC}{Nonlinear Model Predictive Control}
\newacronym{mpc}{MPC}{Model Predictive Control}
\newacronym{panoc}{PANOC}{Proximal Averaged Newton-type method for Optimal Control}

\begin{document}
\def\BibTeX{{\rm B\kern-.05em{\sc i\kern-.025em b}\kern-.08em
    T\kern-.1667em\lower.7ex\hbox{E}\kern-.125emX}}
\markboth{IEEE Transactions on Control Systems Technology, VOL. XX, NO. XX, XXXX 2021}
{Lindqvist \MakeLowercase{\textit{et al.}}: Reactive Navigation of an UAV}

\title{Reactive Navigation of an Unmanned Aerial Vehicle with Perception-based Obstacle Avoidance Constraints}
\author{Bj\"orn Lindqvist, Sina Sharif Mansouri, Jakub Haluška, and George Nikolakopoulos
\thanks{Manuscript received: July 7, 2020. Revised manuscript received: May 21, 2021. Accepted for publication: October 4th 2021.}
\thanks{This work has been partially funded by the European Unions Horizon 2020 Research and Innovation Programme under the Grant Agreement No. 869379 illuMINEation.} 
\thanks{Bj\"orn Lindqvist, Sina Sharif Mansouri, Jakub Haluska, and George Nikolakopoulos are with the Robotics and Artificial Intelligence Team, Department of Computer, Electrical and Space Engineering, Lule\r{a} University of Technology, Lule\r{a} SE-97187, Sweden. Emails of authors in order of appearance: \texttt{bjolin@ltu.se, sinsha@ltu.se, jakhal@ltu.se, geonik@ltu.se.}}
}

\maketitle

\begin{abstract}
In this article we propose a reactive constrained navigation scheme, with embedded obstacles avoidance for an Unmanned Aerial Vehicle (UAV), for enabling navigation in obstacle-dense environments. The proposed navigation architecture is based on Nonlinear Model Predictive Control (NMPC), and utilizes an on-board 2D LiDAR to detect obstacles and translate online the key geometric information of the environment into parametric constraints for the NMPC that constrain the available position-space for the UAV. This article focuses also on the real-world implementation and experimental validation of the proposed reactive navigation scheme, and it is applied in multiple challenging laboratory experiments, where we also conduct comparisons with relevant methods of reactive obstacle avoidance. The solver utilized in the proposed approach is the Optimization Engine (OpEn) and the Proximal Averaged Newton for Optimal Control (PANOC) algorithm, where a penalty method is applied to properly consider obstacles and input constraints during the navigation task. The proposed novel scheme allows for fast solutions, while using limited on-board computational power, that is a required feature for the overall closed loop performance of an UAV and is applied in multiple real-time scenarios. The combination of built-in obstacle avoidance and real-time applicability makes the proposed reactive constrained navigation scheme an elegant framework for UAVs that is able to perform fast nonlinear control, local path-planning and obstacle avoidance, all embedded in the control layer.
\end{abstract}


\glsresetall 

\section{Introduction}\label{sec:intro}

\subsection{Applications and Problem Statement}
\IEEEPARstart{D}{ue} to the latest massive technological improvements in computation power and smart systems, the Unmanned Aerial Vehicles (UAVs) have turned out to be the ultimate all-purpose tool for inspection and exploration tasks. Nowadays, UAVs demonstrate their promising capabilities in application domains, such as wind-turbine inspection~\cite{mansouri2018cooperative} and underground mine navigation~\cite{mansouri2019visioncnn}, search and rescue missions~\cite{tomic2012toward}, including the delivering of first-aid or defibrillators in case of an accident~\cite{zegre2018delivery}, to name a few. The main inspiration for this work was related to subterranean applications in the context of the DARPA Subterranean Challenge~\cite{darpa, IEEEdarpa} as part of Team CoSTAR~\cite{costar} and the Nebula autonomy developments~\cite{agha2021nebula, kim2021plgrim, palieri2020locus}.

Because of the agility of the UAVs, these platforms can access hard-to-reach places, navigate through constrained spaces and ignore any issues related to the ground terrain. The disadvantages of an UAV platform lies in the fast run-time requirements to maintain a flight stability, and the fragility of the platform, where an environmental interaction often leads to a crash. To overcome these problems, effective navigation and obstacle avoidance frameworks are needed with the capability to operate during high run-time requirements, while considering the nonlinear dynamics of the UAV.
 
As UAV's start appearing in more and more applications, the need for stable and intelligent control algorithms is increasing as well. Moving from a human controlled or human aided UAV to a fully autonomous operation, it requires the controller to naturally and in real-time interact with the environment around the UAV, for obstacle avoidance and fast path planning and re-planning. Thus, this article proposes a totally novel reactive local path planning and obstacle avoidance scheme by utilizing a Nonlinear Model Predictive Controller (NMPC) and related environmental information from 2D LiDAR measurements that aims for a fully autonomous online navigation through constrained environments, that can be applied in complex exploration or agile inspection tasks. 
\subsection{Background and Motivation}
The problem of path planning is fundamental in robotics and as such it has been very well studied~\cite{lavalle2006planning}. The UAV brings its own set of challenges to the path planning problem, namely fast run-time requirements, sensitivity to collisions, and nonlinear system dynamics, while many approaches have been successfully applied in real life use cases~\cite{goerzen2010survey}. In general, for the problem of path planning in a constrained environment, the planners can be divided into two generic categories, namely that of reactive online path planners and global path planners, often based on occupancy maps. 

The nominal global planners are the methods based on Dijkstra's algorithm and the Rapidly-exploring Random Tree (RRT) algorithms~\cite{kuffner2000rrt}, with more modern versions, such as Jump-Point Search~\cite{duchoe2014path} and $RRT^x$~\cite{otte2016rrtx} respectively, both capable of a quick re-planning. Despite this, global planners often suffer from a high computation time, drifts in localization and mapping, and generally being dependant on an occupancy map of the surrounding environment. As such, global planners, while necessary for many missions and tasks, require to be paired with a reactive planner, running online and paired with the control layer to guarantee no collisions with the environment. For example, in \cite{pharpatara20163} a RRT-based global planner is paired with an Artificial Potential Field (APF) for UAV navigation. 
Due to multiple benefits, the APF~\cite{rimon1992exact} is the current most common approach utilized for reactive obstacle avoidance and has been used for the UAV application use-cases for both mapping~\cite{droeschel2016multilayered} and exploration tasks~\cite{kanellakis2018towards}, as an extra reactive control layer for obstacle avoidance. 

There are also reactive, or local planning, approaches to occupancy-map based planning~\cite{hrabar2011reactive, oleynikova2016continuous}, where by smart partitioning of the map, the computation time can be greatly reduced. Several methods that link the visual information from depth or monocular cameras to the  optimization problems have also been attempted~\cite{schaub2016reactive, ruf2018real}. 

Another optimization-based method of path-planning is Model Predictive Control (MPC)~\cite{alexis2011switching}, where predicted future states are directly linked to a dynamic model of the system, and as such described by a series of optimized control inputs, acting on the model within the dynamic constraints. With the gaining popularity of Nonlinear Model Predictive Control (NMPC), and with more powerful computation and smarter optimization algorithms, such planners are a perfect fit for the nonlinear and dynamically constrained UAV system. Thus, if obstacle avoidance is included in an NMPC scheme, the result is a reactive planner, that completely considers the dynamics and constraints of the system. 

This approach has gained attention in the latest years. In \cite{rosolia2016autonomous}, a NMPC scheme was developed for an autonomous car with integrated obstacle avoidance of elliptical obstacles, based on the Generalized Minimal Residual (GMRES) method. In \cite{soloperto2019collision} a robust MPC is developed that is in addition also capable of dealing with uncertain moving obstacles of general shapes, evaluated in simulation for an autonomous car.
In \cite{sathya2018embedded}, the Proximal Averaged Newton for Optimal Control (PANOC) algorithm~\cite{stella2017simple} was applied for this purpose, using a ground robot, by considering nonlinear constraints to limit the available position-space of the robot. The benefit of this type of method is the combination of control, local path planning and obstacle avoidance, into one control layer, based on nonlinear optimization. This method has also been applied to an UAV~\cite{small2019aerial}, where the UAV successfully avoided a pre-defined cylindrical obstacle in a laboratory environment, based on nonlinear constraints. In \cite{lindqvist2020nonlinear}, more complex obstacle geometries, and multiple obstacles, are considered for the UAV navigation and a penalty method~\cite{Hermans:IFAC:2018, hermans2021penalty} is applied for the consideration of constraints, while in \cite{sharif2020subterranean} a constrained NMPC was used to keep a safety distance from tunnel walls in a subterranean application. 

Using the Optimization Engine (OpEn)~\cite{open2019, sopasakis2020open} and the PANOC algorithm, this article aims to extend the results of this line of approaches to a fully autonomous and reactive obstacle avoidance and path planning scheme, where on-board 2D LiDAR data is used to form parametric constraints, based on a geometric approximation of the surrounding environment, to guarantee collision-free navigation for multi-obstacle constrained environments, while keeping the solver time low enough for stable control of the system while relying only on limited on-board computation. Extending this obstacle avoidance approach to real-time and  real-life experiments, using on-board sensors for obstacle detection, links the nonlinear dynamics directly to the local path planning and obstacle avoidance problem in a realistic context.

%
\subsection{Contributions}
Based on the aforementioned state of the art, this article focuses on the missing key ingredient in previous NMPC-based obstacle avoidance schemes, namely the fundamental seamless and functional link between the NMPC and the environment awareness, with an extended experimental verification of the overall concept and proper validation of the combined proposed architecture that could stand as a stable and properly functional navigation framework. Towards this direction, the underlying work on the NMPC scheme is a natural continuation in the form of a real-life implementation/experimentation of the theoretical framework proposed in~\cite{sathya2018embedded, small2019aerial, sopasakis2020open}, developed in the context of real application scenarios. This is one of the major contributions of the article, since all previous works, to the authors best knowledge, rely either on simulations~\cite{rosolia2016autonomous, soloperto2019collision, lindqvist2020nonlinear, mansouri2020unified}, or when experiments are performed rely on pre-defined obstacles \cite{small2019aerial, sathya2018embedded}, or motion-capture system to track obstacles \cite{kamel2017nonlinear, lindqvist2020dynamic}. This limits the real-world impact of NMPC-based obstacle avoidance in the robotics context and we show, for the first time, in multiple challenging laboratory experiments the application of an NMPC-based obstacle avoidance scheme in real-time, real-world scenarios for an Unmanned Aerial Vehicle (UAV), relying only on on-board computation and obstacle detection using on-board 2D LiDAR, forming a completely novel navigation scheme for constrained environments that constitutes the second major contribution. In this framework, the NMPC commands roll, pitch, and thrust references at a high frequency and as such merges set-point tracking with the obstacle avoidance problem. As it will be experimentally demonstrated the novel proposed reactive scheme is computationally efficient and fast enough to satisfy the run-time requirements of the UAV platform,  using limited computation power, while maintaining a safe distance from any obstacle and performing smooth and efficient obstacle avoidance maneuvers. 
Additionally we offer a comparison with two artificial potential field avoidance formulations, and we shall show that the NMPC-based avoidance outperforms the artificial potential field in every considered aspect. 
\subsection{Outline} \label{sec:outline}
The rest of this article is structured as follows. Initially, the dynamic model of the UAV is presented in Section~\ref{sec:mavkinematic}, followed by the presentation of the corresponding cost function and the formulation of the obstacle constraints in Sections~\ref{sec:obj} and~\ref{sec:const} respectively. The proposed optimization is presented in Section~\ref{sec:solver}, the two comparison APF formulations are presented in Section \ref{sec:comparison}, while the experimental set-up and tuning is presented in Section \ref{sec:setup}. Multiple experimental scenarios, with related results that display the efficiency of the proposed framework, are presented in Section~\ref{sec:results}, with an additional corresponding comparison and discussion. In Section~\ref{sec:limitations} we discuss the discovered limitation of the framework, offer directions for future works and additions to the method. Finally, Section~\ref{sec:conclusion} concludes this article by summarizing the findings.
\section{Methodology} \label{sec:methodology}

\subsection{System Dynamics} \label{sec:mavkinematic}
\begin{figure}
\centering
  \includegraphics[width=0.9\linewidth]{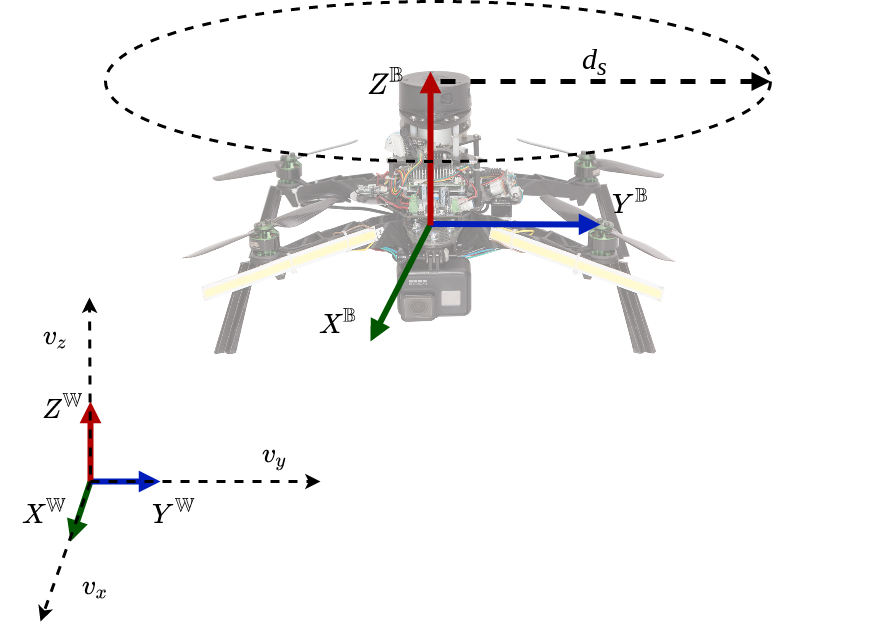}
  \caption{Utilized coordinate frames, where $\mathbb{W}$ and $\mathbb{B}$ denote the world and body coordinate frames respectively. Additionally, the safety distance around the UAV is denoted $d_s$.}
  \label{fig:coordinateMAV}
\end{figure}

%
The UAV coordinate systems are depicted in Figure~\ref{fig:coordinateMAV}, where $(x^\mathbb{B}, y^\mathbb{B}, z^\mathbb{B})$ denote the body-fixed coordinate system, while $(x^\mathbb{W}, y^\mathbb{W}, z^\mathbb{W})$ denote the global coordinate system. In this article, the states are defined in a yaw-compensated global frame of reference.
The six degrees of freedom (DoF) UAV is defined by the set of equations \eqref{eq:mavkinematic}. The same model has been successfully used in previous works, such as in \cite{small2019aerial, lindqvist2020nonlinear, sharif2020subterranean}.  
\begin{subequations}
\small
\label{eq:mavkinematic}
\begin{align}
        \dot{p}(t) &= v(t) \\ 
        \dot{v}(t) &= R(\phi,\theta) 
        \begin{bmatrix} 0 \\ 0 \\ T_{\mathrm{ref}} \end{bmatrix} + 
        \begin{bmatrix} 0 \\ 0 \\ -g \end{bmatrix} - 
        \begin{bmatrix} A_x & 0 & 0 \\ 0 &  A_y & 0 \\ 0 & 0 & A_z \end{bmatrix} v(t), \\ 
        \dot{\phi}(t) & = \nicefrac{1}{\tau_{\phi}} (K_\phi\phi_{\mathrm{ref}}(t)-\phi(t)), \\ 
        \dot{\theta}(t) & = \nicefrac{1}{\tau_{\theta}} (K_\theta\theta_{\mathrm{ref}}(t)-\theta(t)),
\end{align}
\end{subequations}
where $p=[p_x,p_y,p_z]^\top$ is the position, $v = [v_x,v_y,v_z]^\top$ is the linear velocity in the global frame of reference, and $\phi$ and $\theta \in[-\pi,\pi]$ are the roll and pitch angles along the $x^\mathbb{W}$ and $y^\mathbb{W}$ axes respectively. Moreover, $R(\phi(t),\theta(t)) \in \mathrm{SO}(3)$ is a rotation matrix that describes the attitude in Euler form, with $\phi_{\mathrm{ref}}\in \mathbb{R}$, $\theta_{\mathrm{ref}}\in \mathbb{R}$ and $T_{\mathrm{ref}}\geq 0$ to be the references in roll, pitch and total mass-less thrust generated by the four rotors respectively. The above model assumes that the acceleration depends only on the magnitude and angle of the thrust vector, produced by the motors, as well as the linear damping terms $A_x, A_y, A_z \in \mathbb{R}$ and the gravitational acceleration $g$.

The attitude terms are modeled as a first-order system between the attitude (roll/pitch) and the references $\phi_{\mathrm{ref}}\in \mathbb{R}$, $\theta_{\mathrm{ref}}\in \mathbb{R}$, with gains $K_\phi, K_\theta\in\mathbb{R}$ and time constants $\tau_\phi, \tau_\theta \in \mathbb{R}$. The aforementioned terms model the closed-loop behavior of a low-level controller tracking $\phi_{\mathrm{ref}}$ and $\theta_{\mathrm{ref}}$, which also implies that the UAV is equipped with a lower-level attitude controller that takes thrust, roll and pitch commands and provides motor commands for the UAV, such as in~\cite{jackson2016rosflight}.   
%

\subsection{Cost Function} \label{sec:obj}
%
Let the state vector be denoted by $x = [p, v, \phi, \theta]^\top$, and the control action as $u=[T,\phi_{\mathrm{ref}},\theta_{\mathrm{ref}}]^\top$.
The system dynamics of the UAV are discretized with a sampling time $T_s$ using the forward Euler method to obtain
\begin{equation}
\label{eq:prediction}
    x_{k+1} = \zeta(x_k, u_k).
\end{equation}
This discrete model is used as the \textit{prediction model} of the NMPC. This prediction is done with receding horizon e.g., the prediction considers a set number of steps into the future. We denote this as the \textit{prediction horizon}, $N\in\mathbb{N}$, of the NMPC. In some applications, the \textit{control horizon} is distinct from $N$ but in this article we will only consider the case where they are the same, meaning both control inputs and predicted states are computed in the same horizon without loss of generality. By associating a cost to a configuration of states and inputs, at the current time and in the prediction, a nonlinear optimizer can be tasked with finding an optimal set of control actions, defined by the cost minimum of this \textit{cost function}. 

Let $x_{k+j{}\mid{}k}$ denote the predicted state 
at time step $k+j$, produced at the time step $k$. Also denote the control action as $u_{k+j{}\mid{}k}$.
Let the full vectors of predicted states and inputs along $N$ be denoted as $\bm{x}_{k} = (x_{k+j{}\mid{}k})_{j}$ and
$\bm{u}_{k} = (u_{k+j{}\mid{}k})_{j}$.
The controller aims to make the states reach the prescribed set points, while delivering smooth control inputs. To that end, we formulate the following cost function as:
\begin{multline}
\label{eq:costfunction}
J(\bm{x}_{k}, \bm{u}_{k}; u_{k-1\mid k}) = \sum_{j=0}^{N}   \underbrace{\| x_{\mathrm{ref}}-x_{k+j{}\mid{}k}\|_{Q_x}^2}_\text{State cost} 
\\
+   \underbrace{\| u_{\mathrm{ref}}-u_{k+j{}\mid{}k}\|^2_{Q_u}}_\text{Input cost}
+  \underbrace{\| u_{k+j{}\mid{}k}-u_{k+j-1{}\mid{}k} \|^2 _{Q_{\Delta u}}}_\text{Input change cost},
\end{multline}
where $Q_x\in \mathbb{R}^{8\times8}, Q_u, Q_{\Delta u}\in 
\mathbb{R}^{3\times3}$ are symmetric positive definite weight matrices for the
states, inputs and input rates respectively. In \eqref{eq:costfunction}, the 
first term denotes the \textit{state cost}, which penalizes deviating from a 
certain state reference $x_{\mathrm{ref}}$. The second term denotes the 
\textit{input cost} that penalizes a deviation from the steady-state input 
$u_{\mathrm{ref}} = [g, 0, 0]$ i.e. the inputs that describe hovering in place. Finally, to enforce smooth control actions, a third term is added that penalizes changes in successive inputs, the \textit{input change cost}. It should be noted that for the first time step in the prediction, this cost depends on the previous control action $u_{k-1{}\mid{}k} = u_{k-1}$. 
The UAV platform is susceptible to overly aggressive or sudden control actions in the roll and pitch, which can cause unnecessary wobbling or oscillations while translating, especially if it is carrying on-board sensing equipment, whose performance relies on stable and smooth flight behavior. The downside is that the system will be slower, depending on $Q_{\Delta u}$, in reacting to any new information or changing direction.
%
\subsection{Obstacle Definition and Constraints} \label{sec:const}
%
Following the constraint formulation structure for OpEn used in \cite{sathya2018embedded,small2019aerial}, while also keeping the constraints fully parametric so that their positions and size is part of the input fed to the NMPC scheme, we use the function  $[h]_+=\max\{0, h\}$ as described by~\eqref{eq:max}. This allows us to formulate the constraints as equality expressions such that $[h]_+=0$ implies that the constraint is satisfied.
\begin{equation} \label{eq:max}
[h]_+= 
\begin{cases}
0      & \, \text{if } h \le 0 \\
h  &  \,   \text{otherwise}
\end{cases}
\end{equation}
 Equation \eqref{eq:max} is used for expressing a constrained area by choosing \textit{h} as an expression that is positive while violating the constraint, and negative when the constraint holds. Additionally, by utilizing this constraint representation, it allows for the ability of defining more complex geometries by taking the product of multiple such terms, as they are zero where any of the terms are negative. 
 
 For the case of UAV navigation, there are multiple different types of obstacles that can be encountered in the surrounding environment. Being limited to 2D information of the environment, two different obstacle types are included: circles (cylinders) and rectangles (general wall-like obstacles), and their parametric presentation as corresponding constraints will be discussed in the sequel. Moreover, to guarantee bounds on changes in the control actions, a constraint on the consecutive changes in control input will be also established. Finally, it should be highlighted that all the underlying constraints are considered in the full control horizon \textit{N} to account for the obstacle constraints at all the predicted future time steps.

\subsubsection{Circular Obstacle} 
\begin{figure}
\centering
  \includegraphics[width = 0.9\linewidth]{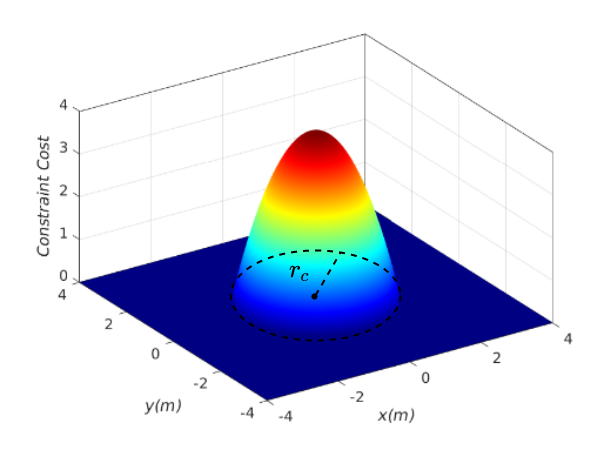}
  \caption{Cost map of cylinder with radius 2m.}
  \label{fig:costplotcylinder}
\end{figure}

A circular constraint can be used for any blocked area or general obstacle, where the radius of the circle envelops the area that is undesirable or blocked, forming an infinite cylinder in the 3D space. The circular constraint is defined in \eqref{eq:cylinderconstraint} with a specified radius and \textit{x-y} position as:
\begin{align}\label{eq:cylinderconstraint} 
    h_{\mathrm{circle}}(p, \xi^{\mathrm{c}}) {}\coloneqq{} [r_{\mathrm{c}}^2 - (p_x{-} p_x^{\mathrm{c}})^2  - (p_y{-}p_y^{\mathrm{c}})^2]_+ = 0,
\end{align}

where $\xi^{\mathrm{c}} = [p_x^{c}, p_y^{c}, r_{c}]$ define the \textit{x,y} coordinates of the center and the radius of the obstacle. 
 A visual representation of the obstacle can be seen in Figure \ref{fig:costplotcylinder}, in the form of a cost map. This is a 2D slice of the 3D space where the new third axis represents the cost related to violating the obstacle constraint. 
\subsubsection{Rectangular Obstacle}\label{sec:line} 
\begin{figure}
\centering
  \includegraphics[width = 0.75\linewidth]{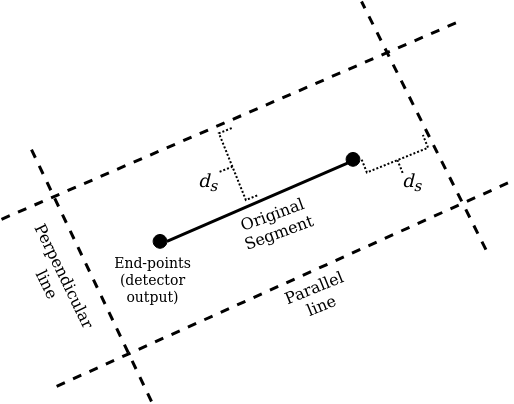}
  \caption{Four lines envelop a line segment, with safety distance $d_s$.}
  \label{fig:rectangle}
\end{figure}

\begin{figure}
\centering
  \includegraphics[width = 0.9\linewidth]{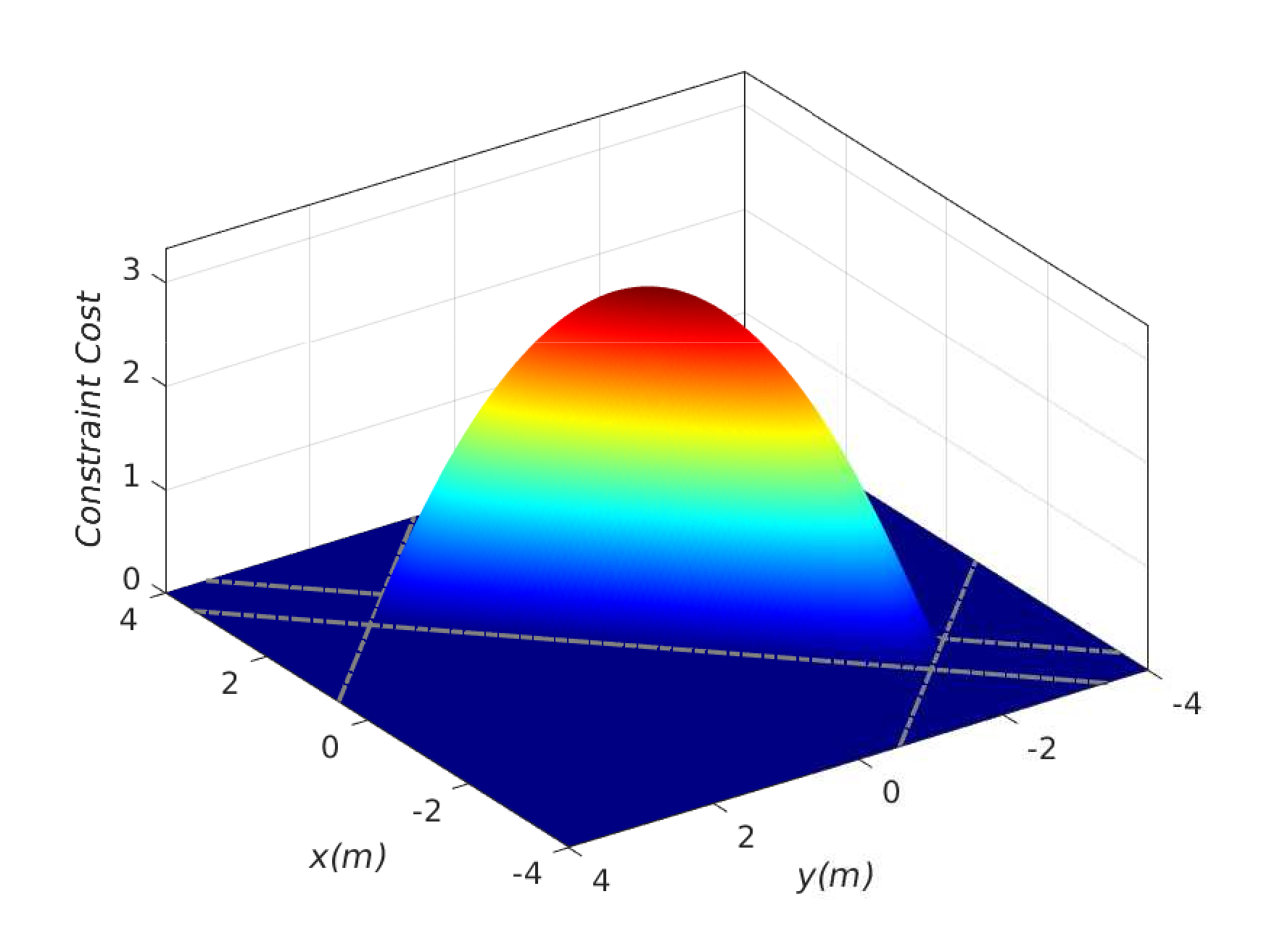}
  \caption{Cost map of a rectangular obstacle as the intersection area of four lines.}
  \label{fig:costplotline}
\end{figure}

The rectangular constraint represents wall-like obstacles with a limited width and length, described by intersecting lines (or hyperplanes) that form a rectangular area. As the NMPC-structure requires that the obstacle structure is pre-defined, we require that the same obstacle-types can be utilized for many situation. Thus, a wall-like obstacle can be defined by the intersection of four lines with safety distance $d_s$ from an original line segment, provided from the environment data (Figure 3). The constraint for a single line takes the form of \eqref{eq:line},
\begin{align}\label{eq:line} 
    h_{\mathrm{line}}(p; \xi^{\mathrm{l}}) {}\coloneqq{} [m p_x - p_y + b]_+ = 0,
\end{align}
where $m, b$ are standard line constants. To describe the rectangular constraint we can take the product of four such terms as: 
\begin{multline}\label{eq:lineconstraint} 
    h_{\mathrm{rec}}(p, \xi^{\mathrm{rec}}) {}\coloneqq{} [m_{\mathrm{par}} p_x - p_y + b_{\mathrm{par,1}}]_+ \\
    [-(m_{\mathrm{par}} p_x - p_y + b_{\mathrm{par,2}})]_+ \\
    [m_{\mathrm{perp}} p_x - p_y + b_{\mathrm{perp,1}}]_+ \\
    [-(m_{\mathrm{perp}} p_x - p_y + b_{\mathrm{perp,2}})]_+ = 0,
\end{multline}
where line constants can easily be computed via simple algebraic operations to form the rectangle in Figure \ref{fig:rectangle} which produces a constraint as in Figure \ref{fig:costplotline}, where $\xi^{\mathrm{rec}} \in \mathbb{R}^6$ includes line constants for all four lines. It should be clear that to form such a constraint, we will have two lines parallel (with constants $m_{\mathrm{par}}, b_{\mathrm{par,1}}, b_{\mathrm{par,2}}$, and opposing signs) to the original line segment and two perpendicular (with constants $m_{\mathrm{perp}}, b_{\mathrm{perp,1}}, b_{\mathrm{perp,2}}$). To reduce large variations in the cost-gradient of lines of varying slopes (especially close-to-1D lines), we pre-condition this constraint using the computed slopes of the lines with the term $1 / (\mid \mid m_{\mathrm{par}}  \mid \mid +  \mid \mid m_{\mathrm{perp}}  \mid \mid)$ for an improved solver consistency.

\subsubsection{Control Input Rate}
We impose a constraint on the successive differences of control actions $\phi_{\mathrm{ref}}$ and $\theta_{\mathrm{ref}}$, so as to directly prevent an overly aggressive behavior of the controller in-flight, that is
\begin{subequations}
\begin{align}
    |\phi_{\mathrm{ref}, k+j-1{}\mid{}k} - \phi_{\mathrm{ref},k+j{}\mid{}k}| 
    {}\leq{}
    \Delta \phi_{\max},
    \\
    |\theta_{\mathrm{ref}, k+j-1{}\mid{}k} - \theta_{\mathrm{ref},k+j{}\mid{}k}| 
    {}\leq{}
    \Delta \theta_{\max}.
\end{align}
\end{subequations}
The above inequality constraints can be re-written as equality constraints as it follows: 
\begin{subequations}\label{eq:delta_constraints}
\begin{align}
    [\phi_{\mathrm{ref}, k+j-1{}\mid{}k} 
    - \phi_{\mathrm{ref},k+j{}\mid{}k}
    -\Delta \phi_{\max}]_+ {}={}& 0,
    \\
    [\phi_{\mathrm{ref},k+j{}\mid{}k}
    -\phi_{\mathrm{ref}, k+j-1{}\mid{}k}
    -\Delta \phi_{\max}]_+ {}={}& 0,
\end{align}
\end{subequations}
e.g. setting a lower and upper bound with the maximum allowed change in consecutive control action as $\Delta \phi_{\mathrm{max}}$. The exact same constraint is also formed for $\theta_{\mathrm{ref}}$, with the maximum change as $\Delta\theta_{\mathrm{max}}$.

\subsubsection{Input constraints} \label{sec:input_constraints}
Finally, we also directly apply constraints on the control inputs. Since the NMPC is used with a real UAV, hard bounds on reference angles $\phi_{\mathrm{ref}}, \theta_{\mathrm{ref}}$ must be considered, as a low-level controller will only be able to stabilize the attitude within a certain range. Since the thrust of a UAV is limited, such hard bounds must also be applied to the thrust input, $T_{\mathrm{ref}}$. Thus, we can define bounds on inputs as:
\begin{equation}
\label{eq:input_const}
u_{\min} \leq u_{k+j\mid k} \leq u_{\max}.
\end{equation}
\subsection{Embedded Optimization} \label{sec:solver}
The NMPC problem is solved by the open-source Optimization Engine (OpEn)~\cite{open2019, sopasakis2020open} and its associated algorithm PANOC~\cite{sathya2018embedded,small2019aerial}, that solves nonlinear non-convex optimization problems. The Optimization Engine generates embedded-ready source code from a specified cost function and set of constraints, while it can solve general parametric optimization problems on the form:
 \begin{subequations}\label{eq:open_problem}
\begin{align}
    \operatorname{Minimize}_{z \in Z}\,& f(z,\rho)
    \\
    \text{subject to:}\,& G(z,\rho) = 0,
\end{align}
\end{subequations}
where $f$ is a continuously differentiable function with Lipschitz-continous gradient function and $G$ is a vector-valued mapping so that $\|G(z,\rho)\|^2$ is a continuously differentiable function with Lipschitz-continous gradient. The decision variable and parameter are denoted by $z$ and $\rho\in \mathbb{R}^{n_p}$ respectively. $\rho$ is an input parameter to the NMPC module including the initial measured state of the UAV $\hat{x}_k = x_{k|k}$, references $u_{ref}$ and $x_{ref}$, the previous control action $u_{k-1}$ (for the first term in the input change cost), and importantly the parametric obstacle data (as described by (5)-(7)), where $n_p$ is the total number of such input parameters.
 
Based on the cost function and constraints, outlined in Sections \ref{sec:obj} and \ref{sec:const}, we can formulate the NMPC problem, while in the presence of $N_c \in \mathbb{N}$ circular obstacles and $N_r \in \mathbb{N}$ rectangular obstacles, as:
 \begin{subequations}\label{eq:nmpc}
\begin{align}
    \operatorname*{Minimize}_{
        \bm{u}_k, \bm{x}_k
    } \,
    & J(\bm{x}_{k}, \bm{u}_{k}, u_{k-1\mid k})
    \\
    \text{s. t.:}\,& 
    x_{k+j+1\mid k} = \zeta(x_{k+j\mid k}, u_{k+j\mid k}),\notag
     \\ & j=0,\ldots, N-1,
    \\
    &u_{\min} \leq u_{k+j\mid k} \leq u_{\max},\, j=0,\ldots, N,
    \\
    &h_{\mathrm{circle}}(p_{k+j\mid k}, \xi_i^{\mathrm{c}}) = 0,\,
     j=0,\ldots, N, \\
     &i = 1,\ldots, N_c
     \\
    &h_{\mathrm{rec}}(p_{k+j\mid k}, \xi_s^{\mathrm{rec}}) = 0,\,
     j=0,\ldots, N, \\
     &s = 1, \ldots, N_r
     \\
     &\text{Constraints \eqref{eq:delta_constraints}},
     j=0,\,\ldots, N, 
     \\
     &x_{k\mid k} {}={} \hat{x}_k.
\end{align}
\end{subequations}

This can be fit into the OpEn framework by performing a \textit{single-shooting} of the cost function via the decision variable $z = \bm{u}_k$ and define $Z$ by the input constraints \eqref{eq:input_const}. We also define $G$ to cast the equality constraints presented in Section \ref{sec:const}. For the consideration of the constraints, a quadratic penalty method\cite{Hermans:IFAC:2018, hermans2021penalty} is applied, as it allows for the types of equality constraints proposed in Section \ref{sec:const}, especially the rectangles formed by products of $[h]_+$ terms, requiring only that the mapping of the constraint expression is continuously differentiable with Lipschitz-continous gradient. By formulating the problem as:
\begin{equation}\label{eq:penaltymethod}
    \operatorname{Minimize}_{z \in Z} f(z,\rho) + q\|G(z,\rho)\|^2,
\end{equation}
where $q \in \mathbb{R}_+$ is a positive penalty parameter, the PANOC algorithm\cite{sathya2018embedded} can be applied to the problem. Using a penalty method, an optimization problem, where the constraints are mapped to the cost-domain, is re-solved multiple times with an increasing penalty parameter $q$ associated to the constraints, while using the previous solution as the initial guess. In very simplified terms, this method gradually moves the cost-minima of \eqref{eq:penaltymethod} by increasing $q$ until non of the constraints are violated, or rather until a specified tolerance is met. 

\begin{figure}[ht]
\centering
  \includegraphics[width=0.8\linewidth]{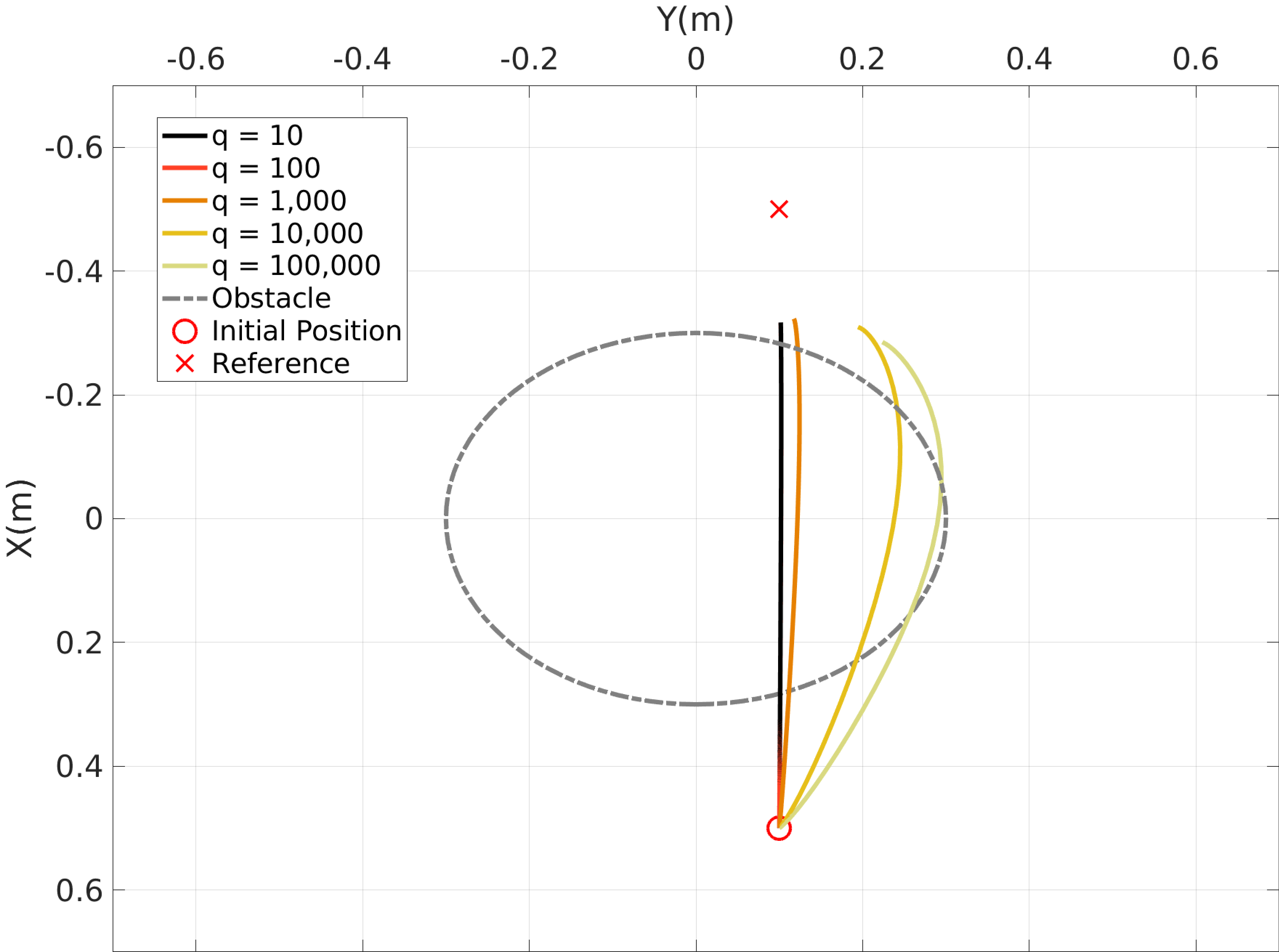}
  \caption{NMPC optimized trajectories using one to five penalty method iterations for a circular obstacle with radius $r = 0.3m$. The cost-minima is gradually pushed out as the cost associated with violating the constraint is increased.}
  \label{fig:penaltyplot}
\end{figure}

In Figure \ref{fig:penaltyplot} the penalty method concept is displayed, where the solution from one to five penalty method iteration are shown for obstacle avoidance around a circular obstacle where $q^i = 10^{i}, i = 1, \ldots, 5$. As the penalty parameter is increased, the optimized trajectory, $\bm{u}_{k}$ (here displayed in position coordinates via the prediction model), is moved out of the obstacle. The trajectory will lie completely outside the obstacle as $q$ approaches infinity, and as such small constraint violations should be expected and compensated by the safety distance $d_s$. In Figure \ref{fig:penaltyplot} it is depicted that the first iterations are not significantly different, and in Section \ref{sec:results} we will be using penalties with $q^i= 10^3\times(4^{i-1}), i = 1, \ldots, 4$, based on the results from initial testing and simulations. 

\subsection{Comparison Methods}\label{sec:comparison}
For performing a comparison with the proposed NMPC approach we will implement two Artificial Potential Field (APF) formulations, based on an approach very similar to the legacy APF by Warren~\cite{warren1989global}, as APFs are one of the most common reactive avoidance formulations and fit very well into the 2D-LiDAR equipped UAV. The APF algorithms have the advantage of being able to work directly with the LiDAR scan 2D pointcloud, producing repulsive forces that shift the translational speed of the UAV based on the proximity and number of points within a certain radius of influence of the forces as $r_F \in \mathbb{R}$ of the APF. The APF should be seen as a pure reactive avoidance layer and requires a separate reference tracking controller, and to keep the comparison fair we will use a similar NMPC, but without integrated obstacle avoidance. In the following sections (\ref{sec:baseline}, \ref{sec:enhanced}) we will define two formulations on the repulsive forces, one closely following the legacy approach, called the baseline APF, and one where we have added enhanced concepts that facilitate problems specific to the UAV platform in densely occupied environments. In regards to the selected optimization software and method for solving obstacle avoidance constraints, detailed comparisons with other optimizers/methods on the computational efficiency and ability to handle constraints of this type are offered in \cite{open2019, hermans2021penalty}.

\subsubsection{Baseline Approach}\label{sec:baseline}
Let us define the LiDAR 2D pointcloud as an array of points \{P\}, where each point is described as the relative \textit{x,y}-position from the UAV/LiDAR (e.g. in the UAV body frame) as $\varrho = [\varrho_x, \varrho_y]$. Let us also denote the repulsive force as $F^r = [F^r_x, F^r_y]$ and the attractive force as $F^a = [F^a_x, F^a_y]$. As we are only interested in points inside the radius of influence, when considering the repulsive force, let's denote the list of such points $\boldsymbol{\varrho}_F \in \{P\}$ where $\mid\mid \varrho^i_F \mid \mid \leq r_F$ and $i = 1,2, \ldots, N_{\varrho_F}$ (and as such $N_{\varrho_F} \in \mathbb{N}$ is the number of points to be considered for the repulsive force). 
We can define the repulsive force as:
\begin{equation}\label{eq:repulsive}
    F^r = \sum^{N_{\varrho_F}}_{i=1}  L^r(1 -  \frac{\mid\mid \varrho_F^i \mid\mid}{r_F})\frac{-\varrho_F^i }{\mid\mid \varrho_F^i \mid\mid} + L^{offset}\frac{-\varrho_F^i }{\mid\mid \varrho_F^i \mid\mid}
\end{equation}
where $L^r = [L^r_x, L^r_y] \in \mathbb{R}^2$ are the repulsive constants/gains representing the maximum repulsive force per point, and $L^{offset} \in \mathbb{R}$ is an additional static potential. We can further define the attractive force to simply be $L^a(p_{ref} - \hat{p})$, where $L^a \in \mathbb{R}$ is the attractive gain, $p_{\mathrm{ref}}$ are the first two elements in $x_{\mathrm{ref}}$ related to the \textit{x,y}-position references, and $\hat{p} \in \mathbb{R}^2$ is the measured \textit{x,y}-position. The total force can then be obtained as $F = F^a + F^r$. To fit the APF to the reference tracking controller, the attractive force can be seen simply as the controller trying to reach the desired way-point reference, while the repulsive force is the momentary shift in that way-point as to avoid any obstacles.

\subsubsection{Enhanced Approach}\label{sec:enhanced}
As we have discussed previously in Sections \ref{sec:obj} and \ref{sec:const}, the UAV as a platform is susceptible to overly aggressive maneuvering, which in very tight spaces can lead to unwanted behavior. As such, we will impose a very similar repulsive force function, but add saturation limits on the magnitude of forces, saturation limits on the rate of change of repulsive forces from one time step to the next, and a normalization of the attractive and summed total forces. Additionally, instead of a static force per point inside $r_F$ we propose a larger static force for points inside a safety-critical radius $r_s \in \mathbb{R}$. We define the set of points, inside the safety-critical radius, as $\boldsymbol{\varrho}_s \in \{P\}$ where $\mid\mid \varrho^j_s \mid \mid \leq r_s$ and $j = 1,2, \ldots, N_{\varrho_s}$ and $N_{\varrho_s} \in \mathbb{N}$ is the number of points inside the safety-critical radius. The advantages, as compared to the baseline approach, are as follows: reduces excessive actuation by too rapid changes in the repulsive force, force normalization that leads to a much more consistent behavior of the reference-tracking controller, while making the tuning process much easier, and the addition of an extra safety-bound if there are any points inside critical radius $r_s$. Thus, we can define the repulsive force as:

\begin{equation}\label{eq:repulsive_b}
    F^r = \sum^{N_{\varrho_F}}_{i=1}  L^r(1 -  \frac{\mid\mid \varrho_F^i \mid\mid}{r_F})^2\frac{-\varrho_F^i }{\mid\mid \varrho_F^i \mid\mid} + \sum^{N_{\varrho_s}}_{j=1} L^{s}\frac{-\varrho_s^j }{\mid\mid \varrho_s^j \mid\mid}
\end{equation}
with $L^s \in \mathbb{R}$ as the large repulsive gain to ensure that the UAV directly moves away from any point inside $r_s$. Imposing the suggested improvements (saturations, normalization) can most easily be explained in the depicted form in Algorithm 1, where $F_{max}\in \mathbb{R}_+$ and $\Delta F_{max}\in \mathbb{R}_+$ are saturation limits on the magnitude and rate of change on the repulsive force, and $F^r_{k}$ and $F^r_{k-1}$ are the current and previous repulsive forces respectively, and $sgn()$ is the sign function.

\begin{algorithm}[htbp]
\SetAlgoLined
\textbf{Inputs:} $F^a, F^r_k, F^r_{k-1}$  \\
\If{$\mid\mid F^r_k \mid\mid > F_{max}$}{ 
        $F^r_k \gets sgn(F^r_k)F_{max}$} 
\If{$\mid\mid F_k - F_{k-1}\mid\mid > \Delta F_{max}$}{ 
        $F^r_k \gets F^r_{k-1} + sgn(F_k - F_{k-1})\Delta F_{max}$ 
}
\If{$\mid\mid F^a \mid\mid > 1$}{
    $F^a \gets \frac{F^a}{\mid\mid F^a \mid\mid}$}
$F \gets F^r_k + F^a$ \\
\If{$\mid\mid F \mid\mid > 1$}{
    $F \gets \frac{F}{\mid\mid F \mid\mid}$}
\textbf{Output:} $F$  
\caption{Force calculation}\label{alg:force_calc}
\end{algorithm}

\subsubsection{Potential Field Tuning}
The APF is extremely dependent on the tuning of repulsive and attractive gains $L^a, L^r$, as well as on the radius of influence $r_F$. Its performance is also very dependant on the tuning of the reference tracking controller that it is paired with, and performance can be reduced from moving aggressively as it leads to more abrupt changes in the repulsive forces. The APFs are evaluated in Section \ref{sec:results} and in three experimental scenarios, where two require the UAV to pass in-between two obstacles. For a fair comparison, we will require that only one tuning of the APF is used for all three experiments to better evaluate the general performance. Based on the authors' experience, for the UAV case the APF generally performs better when $r_F$ can be chosen relatively large, while the UAV is moving relatively slowly as to make the reaction to changing forces less abrupt. However, to match the challenging scenarios that is used for evaluating the NMPC, we are forced to chose a small enough $r_F$ to allow the UAV to move/fit through the densely occupied environment. As such, the force gains are chosen as $L^a = 1, L^r = [0.08, 0.16], L^{offset} = 0.04, L^s = 1.5$, while $r_F = \unit[0.75]{m}$ and $r_s = \unit[0.4]{m}$ (slightly above the UAV size-radius of \unit[0.3]{m}), resulting in a comparatively aggressive tuning but with a smaller radius of influence, while we tune the reference tracking controller to be as fast as possible, while maintaining no collisions. Additionally, saturation limits for the enhanced approach are set to $F_{max} = 6, \Delta F_{max} = 0.5$ respectively.


\section{Experimental Set-up}\label{sec:setup}
\begin{figure*}[ht]
\centering
  \includegraphics[scale=0.36]{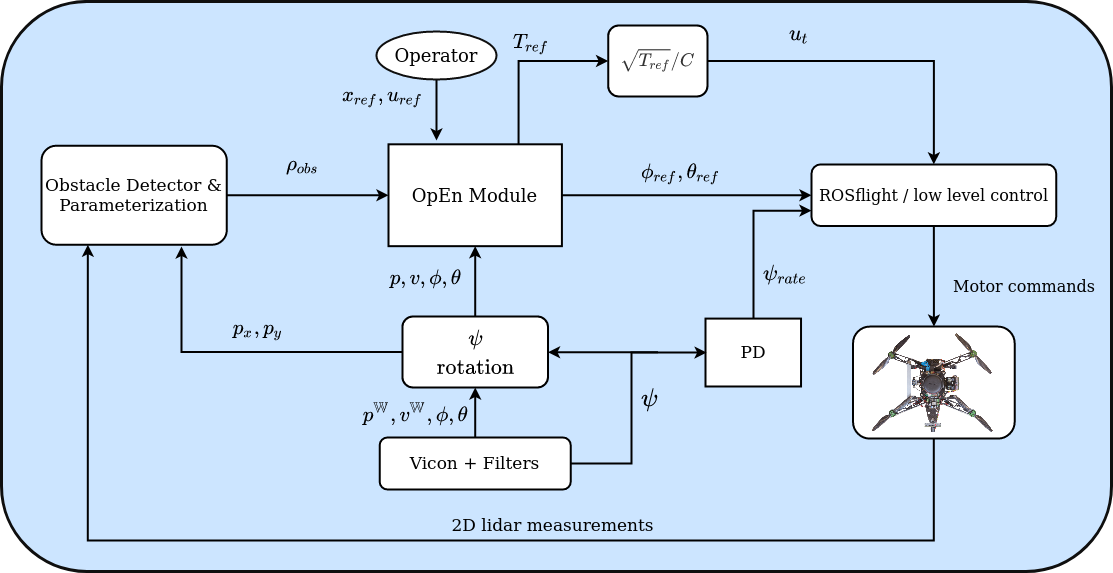}
  \caption{Overall proposed control architecture overview. State estimation is done via a Vicon motion capture sytem and a median filter to estimate velocities. The 2D LiDAR measurements are provided by an on-board RPLiDAR-A3. The Optimization Engine (OpEN) module takes references ($x_{ref}, u_{ref}$), state measurements ($p,v,\theta, \phi$), and obstacle parameters $\rho_{obs}$, to compute the control inputs ($T_{ref}, \theta_{ref}, \phi_{ref}$). These are provided together with a yaw rate command $\psi_{rate}$ (from a decoupled PD-controller) to the ROSflight low level controller that calculates motor commands for the UAV.}
  \label{fig:controlscheme}
\end{figure*}

For the laboratory experiments we use the ROS~\cite{quigley2009ros} (Robot Operating System) architecture for message handling between nodes. The general control structure can be seen in Figure \ref{fig:controlscheme}, where geometric environment data, registered from 2D LiDAR scans, as well as state information ($p,v,\theta, \phi$) from a Vicon motion capture system is fed to the NMPC module, together with references ($x_{\mathrm{ref}}$,$u_{\mathrm{ref}}$ provided by the operator), to compute the control inputs ($T_{ref}, \theta_{ref}, \phi_{ref}$). The utilized low-level controller is ROSflight~\cite{jackson2016rosflight}, which takes roll, pitch, thrust and yaw-rate commands. The thrust command signal $u_t \in [0,1]$ is assumed to have a quadratic relationship to the mass-less thrust (acceleration) input $T_{\mathrm{ref}}$ as:
\begin{equation}\label{eq:thrusteq}
    T_{\mathrm{ref}} = Cu_t^2,
\end{equation}
where $C\in\mathbb{R}$ is the thrust constant that maps between $T_{\mathrm{ref}}$ and $u_t$.
 \subsection{Platform}
\begin{figure}[ht]
\begin{subfigure}{.5\linewidth}
  \centering
  \includegraphics[width=1.2\linewidth]{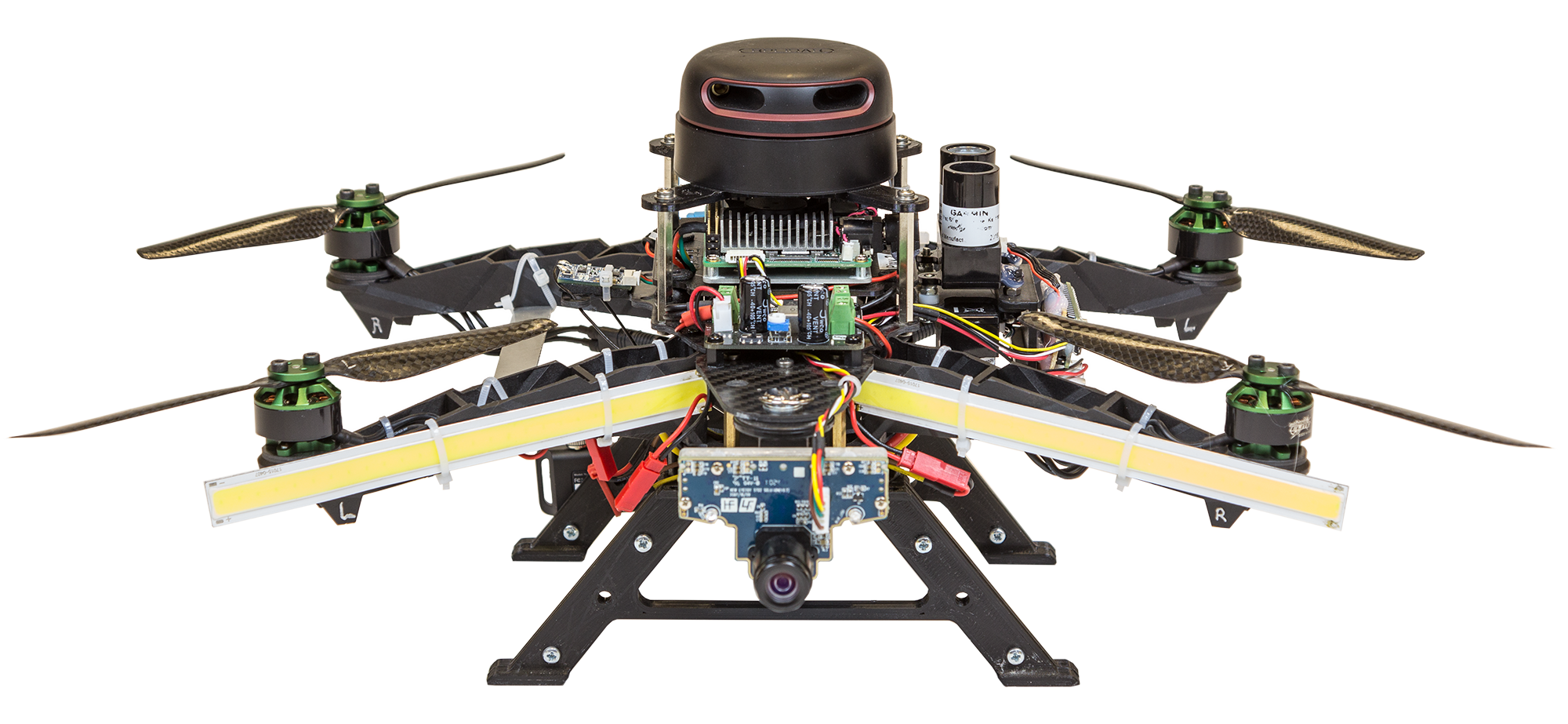}
  \caption{front view}
  \label{fig:sfig1}
\end{subfigure}%
\begin{subfigure}{.5\linewidth}
  \centering
  \includegraphics[width=.7\linewidth]{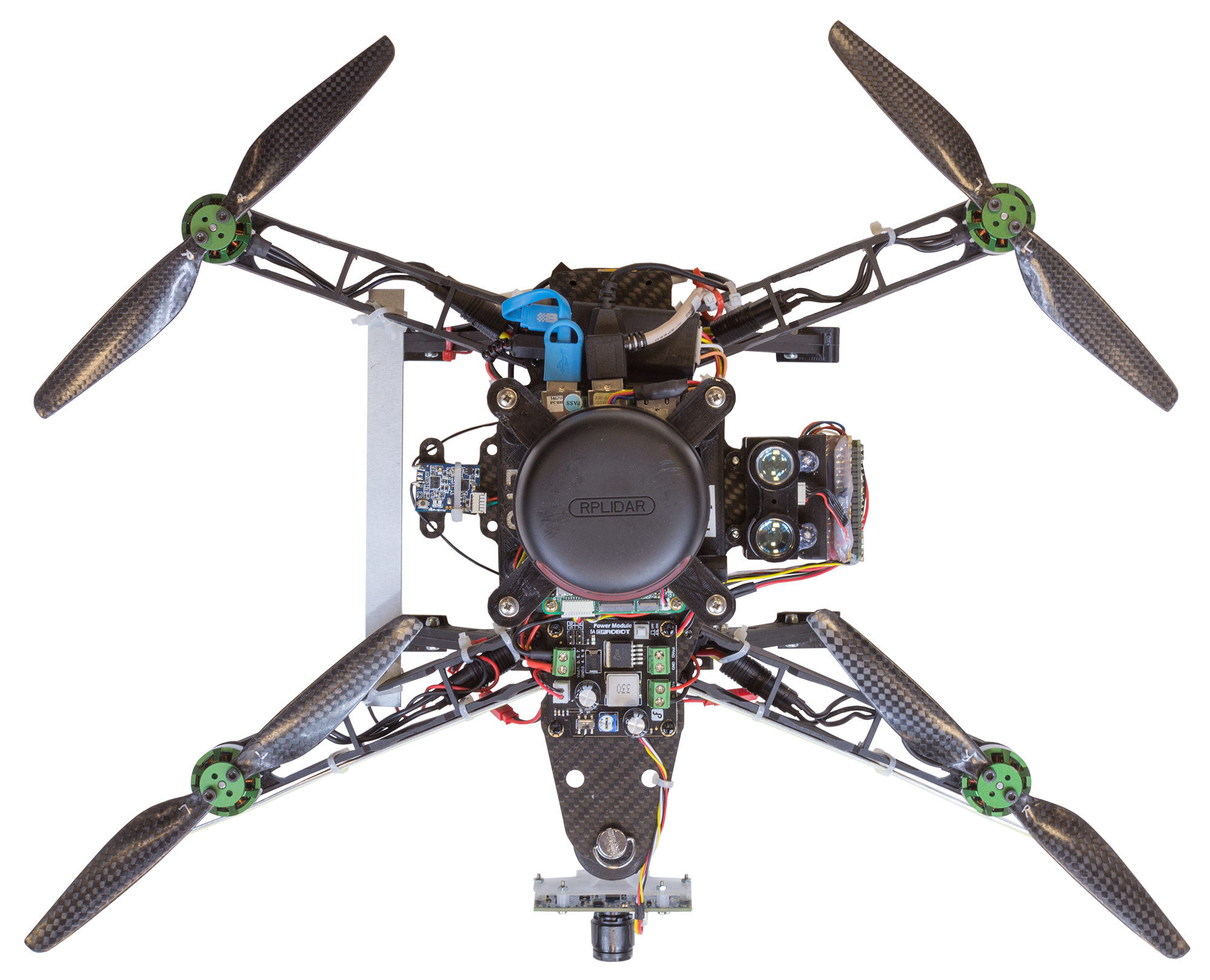}
  \caption{top view}
  \label{fig:sfig2}
\end{subfigure}
\caption{UAV utilized for laboratory experiments. The subterranean scout, named the Pixy.}
\label{fig:pixy}
\end{figure}

The platform used for performing laboratory experiments is the Pixy~\cite{kominiak2020mav} designed at Lule\r{a} University of Technology, seen with its full sensor suite in Figure \ref{fig:pixy}, designed and used for underground constrained environments. The Pixy has been the testbed platform for multiple underground applications~\cite{mansouri2019vision, mansouri2020deploying, kanellakis2018towards}, designed to be a light and low-cost platform for aerial scouting purposes. In these experiments, we will use a simplified model, as state measurements is provided by the motion capture system so no on-board state estimation (except IMU) is required. As such, the important components of the platform are the RPLiDAR-A3 2D LiDAR and the on-board computer, which is an Aaeon UP-Board with an Intel Atom x5-Z8350 processor and 4GB RAM. As such, the NMPC optimization, as well as the 2D LiDAR information processing, is done completely on-board with the limited computation power of the UP-board. 
 
 \subsection{Obstacle Detection}
\begin{figure}[ht]
\centering
  \includegraphics[width=0.8\linewidth]{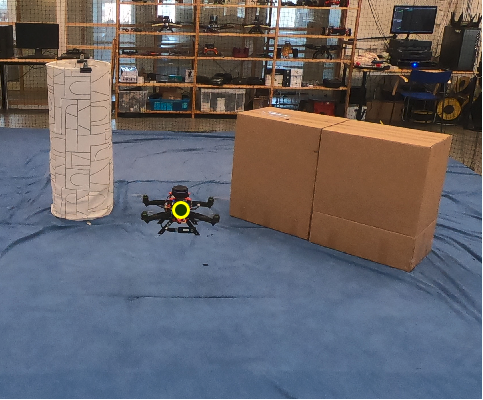}
  \caption{Example scenario of UAV and obstacles in the laboratory environment.}
  \label{fig:drone_obs}
\end{figure}

\begin{figure}[ht]
\centering
  \includegraphics[width=0.8\linewidth]{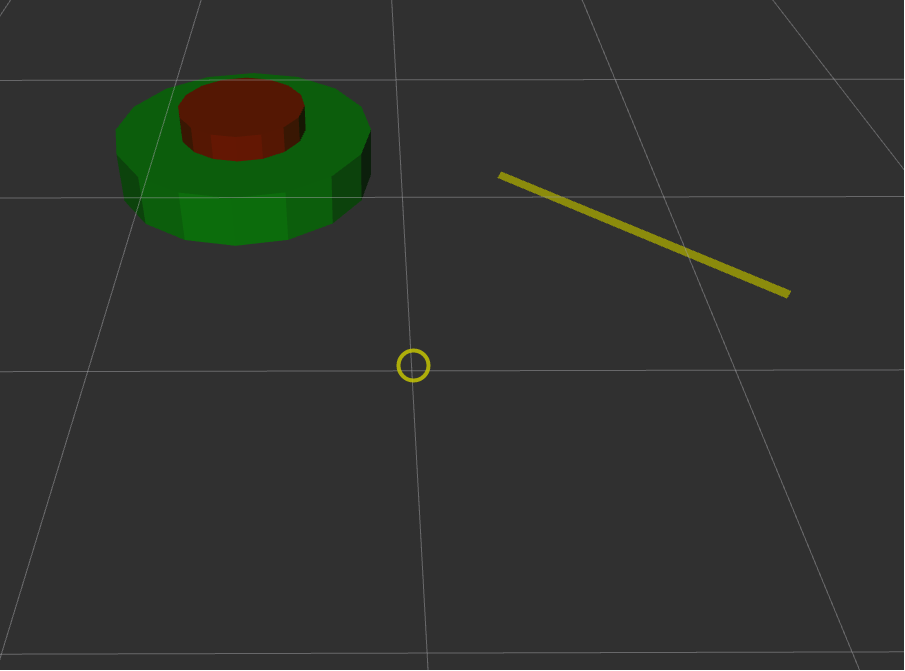}
  \caption{Output from the Obstacle Detector in rviz from the scenario shown in Figure \ref{fig:drone_obs}, showing the radius and safety radius of the circular obstacle and the line segment from the wall-type obstacle.}
  \label{fig:obs_vis}
\end{figure}

\begin{figure}[ht]
\centering
  \includegraphics[width=0.8\linewidth]{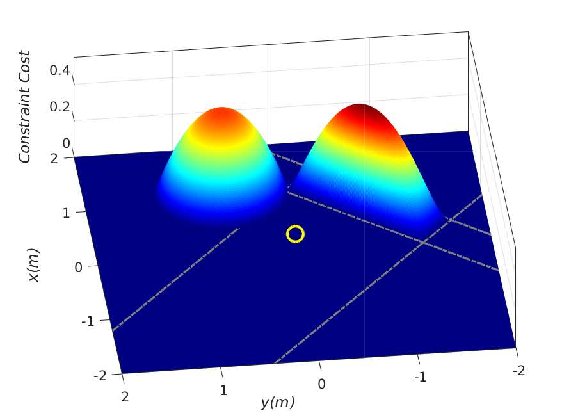}
  \caption{Corresponding costmap from constraints formed with obstacle data from Figure \ref{fig:obs_vis}.}
  \label{fig:cost_vis}
\end{figure}

 Using the 2D LiDAR measurements from the on-board LiDAR, we use the open-source ROS package \textit{Obstacle Detector}~\cite{przybyla2017detection} to detect and track obstacles. The package uses combined segmentation and merging of obstacles from 2D LiDAR data and it is a perfect fit for the required needs, providing geometric approximations of the surrounding environment in the form of line segments and circles. 
 Based on the on-board sensor, only the section of the environment within line-of-sight of the UAV is visible. The obstacle detector matches the visible section of the obstacle to the best-fitting circle segment or line segment. Assuming the obstacle to be circular, with only one side visible, it could lead to incorrect path generation, but as the detector and the NMPC are run at a high frequencies, both the obstacle data and the generated paths are continually updated based on the new upcoming information (for example seeing more of the obstacle as the UAV moves past it), which mitigates such problems.
 Circular obstacles, \textit{c}, are, just like in the constraint formulation, defined by a radius and the \textit{x-y} coordinates of the center of the circle as;
  \begin{equation}
    c \triangleq \{r_c, p^c\} = \{r_c, (p^c_x, p^c_y)\}. 
 \end{equation}
where $r_c, p^c_x, p^c_y$ define the radius, and \textit{x,y}-position of the center of the circle. The obstacle detector also supports an additional safety distance, $d_s$, such that $r_c = r_{real} + d_s$. This safety distance is required, since the constraints for obstacle avoidance assumes the position of the UAV to be expressed by a point. As such, $d_s$ represents the size of the UAV, and also in practise an extra increase to compensate for inaccuracies in measurements, solver tolerances, and limited penalty method iterations. Line segments, \textit{l}, are defined by extreme points of the line segment as:
 \begin{equation}
    l \triangleq \{p^{l,1}, p^{l,2}\} = \{(p_x^{l,1}, p_y^{l,1}), (p_x^{l,2}, p_y^{l,2})\},
 \end{equation}
 where $(p_x^{l,1}, p_y^{l,1})$ defines the \textit{x,y}-position of the start-point of the line segment and $(p_x^{l,2}, p_y^{l,2})$ the end-point.
 From each such set of points, we can parameterize a rectangular constraint as defined in Section \ref{sec:line}. This is done by computing the line equations, as shown in Figure \ref{fig:rectangle} - \ref{fig:costplotline}, to envelop the original line segment with a rectangle using the same extra safety distance $d_s$. 
 As the detector is limited to outputting circles and line segments, obstacles are forced to be classified into either one, or divided into multiple obstacles, as it would be the case for a multiple-sided obstacle. The proposed architecture does not further merge obstacles, while the overlap of the obstacles poses no problems for the optimization method, except by unnecessarily using the limited number of constraints. An example of how the Obstacle Detector approximates the environment and how the constraints are formed can be seen in Figures \ref{fig:drone_obs}-\ref{fig:cost_vis}.
 Although being limited to only considering lines and circles, this method of obstacle detection performs very well and runs in real time and without a loss of generality, which is required for these type of applications. It should also be noted that, for the purpose of obstacle avoidance, only obstacles at a specified distance from the UAV, based on the horizon $N$, need to be considered in the constraint formulation. This greatly reduces the total number of obstacles and as such reduces the overall computational load.  
 
 \subsection{Model Identification}\label{sec:identification}
 In accordance to the model description presented in Section \ref{sec:mavkinematic}, there are a set of unknown model parameters. Since the point of an NMPC scheme is to predict future states based on the optimized trajectory, the better model fit, the better the performance of the NMPC. Thus, the most impactful parameters are the first-order constants, describing the behavior of the UAV when a control input $(\theta_{ref},\phi_{ref})$ is applied. These constants are gains $K_\phi, K_\theta\in\mathbb{R}$ and time constants $\tau_\phi, \tau_\theta \in \mathbb{R}$. These terms are dependant on both the platform and the tuning of the low-level controller and motor mixer, which in this case is the ROSflight. 
 
\begin{figure}
\begin{subfigure}{.5\linewidth}
  \centering
  \includegraphics[width=1\linewidth]{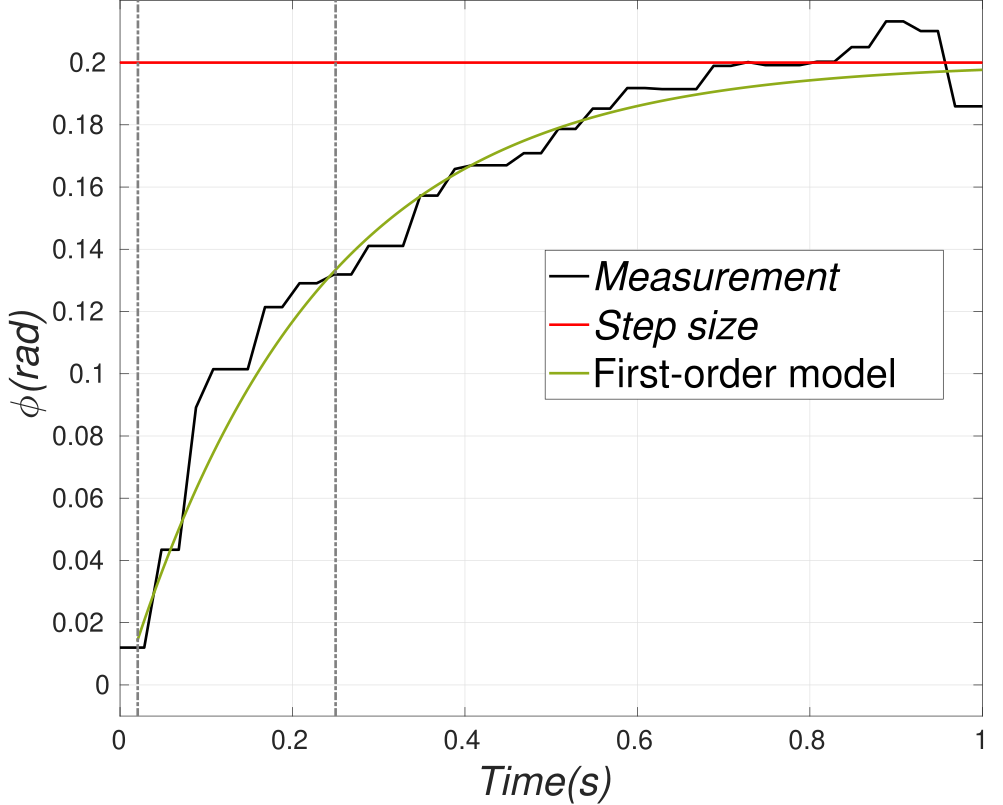}
  \caption{$\phi$ step response}
  \label{fig:rollstep}
\end{subfigure}%
\begin{subfigure}{.5\linewidth}
  \centering
  \includegraphics[width=1\linewidth]{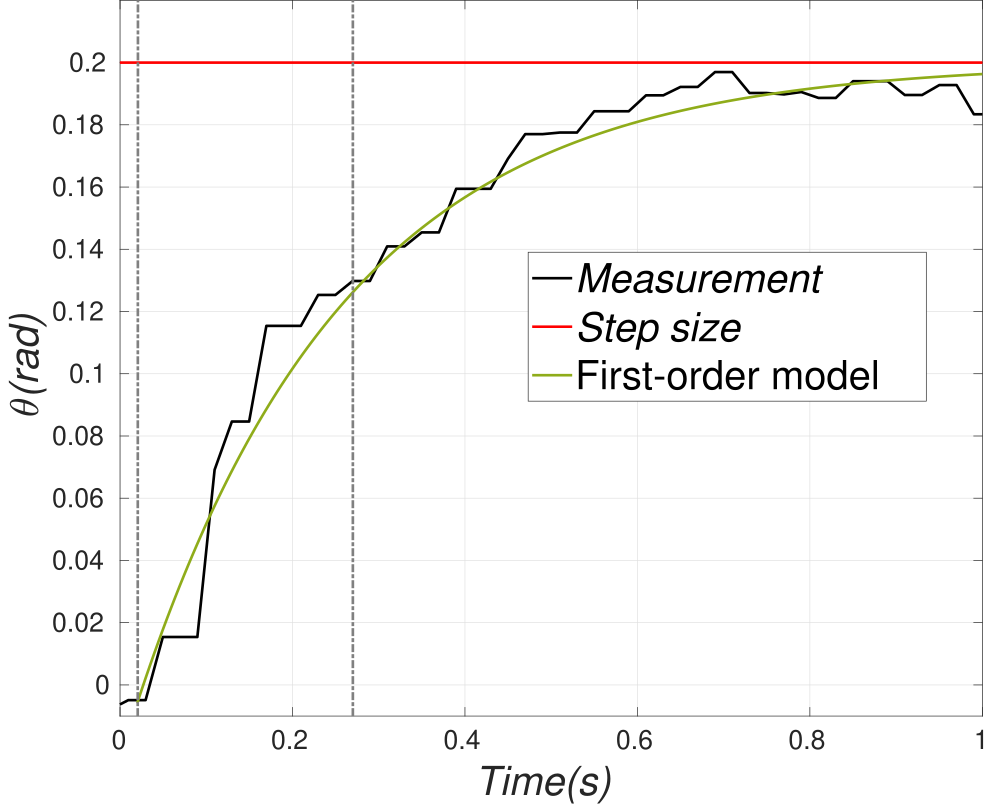}
  \caption{$\theta$ step}
  \label{fig:pitchstep}
\end{subfigure}
\caption{System identification for first-order time constants of LTU-Pixy.}
\label{fig:step}
\end{figure}

 We evaluate these constants by applying a step input in $\theta_{\mathrm{ref}}$ and $\phi_{\mathrm{ref}}$ to the attitude controller (ROSflight) on the UAV during flight and analysing the corresponding response, as depicted in Figure \ref{fig:step}. From the obtained results, it is evident that while not matching perfectly to that of a first order system, we can approximate the time constants $\tau_\phi$ and $\tau_\theta$ to 0.23 and 0.25 respectively, while the gains $K_\phi$ and $K_\theta$ are (close to) unity. Similarly, we perform a height control test to identify the thrust constant $C$, as described in \eqref{eq:thrusteq}, and thus this has been identified to be $\frac{\sqrt{g}}{0.48}$ for a fully charged battery. In-flight we add a weak integrator based on the height measurement to this constant to compensate for slight variations from battery drainage. 

\subsection{NMPC Tuning parameters}\label{sec:tuning}
For the experimental validation of the method, in addition to the model parameters tested for in Section \ref{sec:identification}, we set linear damping terms $A_x, A_y, A_z$ to $0.1, 0.1, 0.2s^{-1}$ respectively. The penalty method parameters are set as described in Section \ref{sec:solver}, while we use a discretization of the cost function with a sampling time of $50ms$ and a prediction horizon $N$ of 40, implying a prediction of two seconds. And as such, we will be running the main control loop at \unit[20]{Hz} as well, which, with the on-board IMU and ROSflight running at \unit[100]{Hz}, is appropriate per common inner/outer control loop dynamics.
To avoid run-time issues we also impose a hard bound on the solver time to \unit[40]{ms}, which means if the solver does not converge in that time, we will use the non-converged solution. The symmetric weight matrices in \eqref{eq:costfunction} are selected as: $Q_x = \operatorname{diag}(2,2,40,5,5,5,8,8)$, $Q_u =\operatorname{diag}(5, 10, 10)$ and $Q_{\Delta u} = \operatorname{diag}(10, 20, 20)$.
%
%
The constraints on control inputs are selected (in SI-units) as: $u_{min} = [5, -0.2, -0.2]^T$ and $u_{max} = [13.5, 0.2, 0.2]^T$

The constraints on the change of the input, described in (\ref{eq:delta_constraints}), are selected as $\Delta\phi_{max} = 0.08$ and $\Delta\theta_{max} = 0.08$, and $d_s$ is set to \unit[0.4]{m} (UAV radius with propellers included ca. \unit[0.3]{m}, and an additional \unit[0.1]{m} for safety). This tuning is extremely conservative with high weights on the inputs and low weights on the position states, which is what we will use for the set-point tracking. Additionally, the input constraints only allow for a small magnitude in $\phi_{\mathrm{ref}}, \theta_{\mathrm{ref}}$. 

This conservative tuning is selected for two reasons: a) low \textit{x,y-}position weights lead to a smaller cost-increase from deviating from the state reference to perform obstacle avoidance maneuvers, b) the high input weights in addition to input constraints keeps the UAV closer to the steady state input of $[9.81,0,0]$ in-flight and as such keeps the 2D LiDAR more stable, which was key to increase the performance of the Obstacle Detector. The trade-off is a decrease in the distance covered inside the prediction horizon (as the UAV will be predicted to move slower). Such a conservative tuning might also be closer to that of a field-application, as the  on-board state estimation often suffers from too quick or sudden maneuvering. 

Finally, we consider obstacles within a $3m$ radius of the UAV in the optimization problem, since by the conservative tuning this is slightly above the maximum distance covered within the prediction horizon. As stated in the optimization problem, the total number of obstacles have to be pre-defined, and we set this number to $N_c = 5$ and $N_r = 10$, which are also sorted by distance to the UAV. The average computation time naturally increases with more defined obstacle constraints and these numbers were chosen as a compromise between computation time and ensuring that all nearby detected obstacles can be included in the NMPC problem.
We did not notice any case where this limitation caused any collisions due to not considering a certain obstacle. 
%

\section{Result} \label{sec:results}
The method for the evaluation of the proposed method will be for the UAV to take off to a set-point reference and will then be given a set-point to go through various scenarios with obstacles or obstacle courses. We kindly recommend the reader to watch the corresponding video of the experiments, as it demonstrates the set-up and the overall performance clearly. Video Link:
\scriptsize{\url{https://youtu.be/xl_YQuDjs1M}}. \normalsize

\begin{figure}[ht]
\centering
  \includegraphics[width=0.9\linewidth]{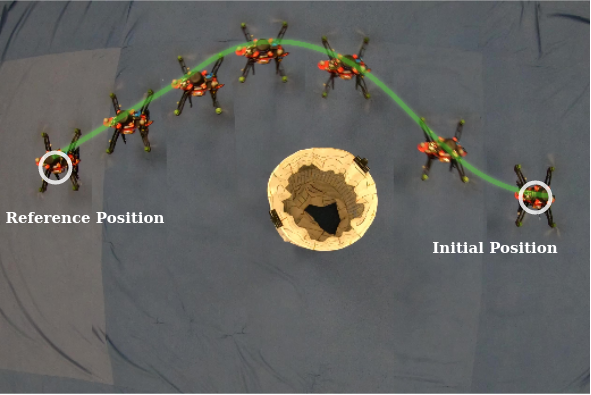}
  \caption{Avoidance of a circular obstacle.}
  \label{fig:cylinder_path}
\end{figure}

\begin{figure}[ht]
\centering
  \includegraphics[width=0.9\linewidth]{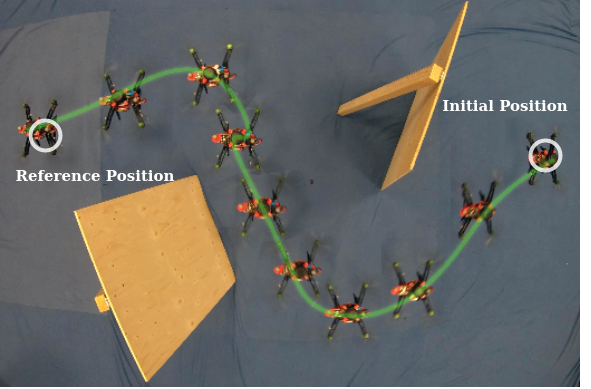}
  \caption{Avoidance of two wall-like obstacles.}
  \label{fig:2wall_path}
\end{figure}

\begin{figure}[ht]
\centering
  \includegraphics[width=0.9\linewidth]{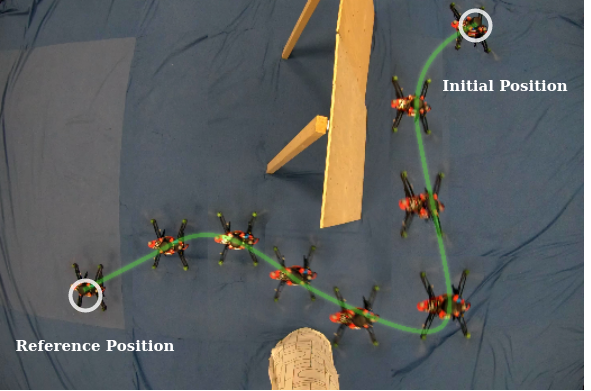}
  \caption{Obstacle avoidance through a small opening.}
  \label{fig:opening_path}
\end{figure}

Figures \ref{fig:cylinder_path}-\ref{fig:opening_path} show the scenarios set up for the experimental validation, as well as the path of the UAV through the constrained environment, while Figures \ref{fig:exp1_merged}-\ref{fig:exp3_merged} show snapshots from the above mentioned video that also include visualization of obstacle representations and the predicted optimized trajectory. Generally, we would like to evaluate three things: a) the various obstacle types, b) the capability of dealing with multiple obstacle, and c) the efficacy of the method at passing through tight openings or passages, which are tested in the three scenarios respectively. 

\begin{figure}[ht]
\centering
  \includegraphics[width=0.8\linewidth]{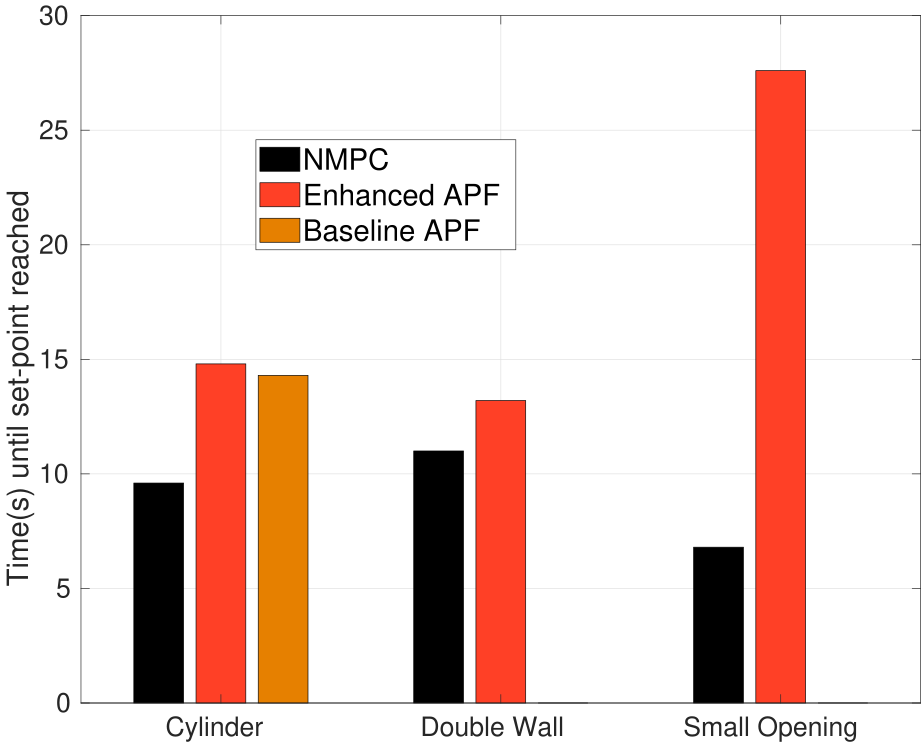}
  \caption{Time duration from initiated movement until reaching the reference for NMPC and APF comparison.}
  \label{fig:setpoint_time}
\end{figure}

As for the comparison with the APF, we will also evaluate three things: a) the time until the reference set-point is reached, b) the efficiency of the avoidance maneuvers, and c) the ability to maintain the desired safety distance of $\unit[0.4]{m}$. The time until mission completion is shown in Figure \ref{fig:setpoint_time}, while a video of the APF performance is shown in \scriptsize{\url{https://youtu.be/dLnoLcNxPzs}}. \normalsize Figures \ref{fig:pf_w_merged}-\ref{fig:pf_b_merged} show the paths through the environment for each APF. From the obtained results, it is clear that the NMPC, by imposing hard bounds and considering a \unit[2]{s} prediction on how to move around the obstacles, greatly outperforms the APF. While the enhanced APF is capable of completing all three scenarios (the baseline approach was simply incapable of moving in-between obstacles without excessive and dangerous movements), the NMPC objectively outperforms it on the time to reach the reference, and looking at the video results or Figures \ref{fig:pf_w_merged}-\ref{fig:pf_b_merged} as compared to Figures \ref{fig:cylinder_path}-\ref{fig:opening_path}, also very greatly outperforms it in the ability to execute efficient avoidance maneuvers. This can greatly be attributed to the \textit{proactive} component of NMPC-based obstacle avoidance, where the UAV can start its avoidance maneuver, as soon as the obstacle is within the prediction horizon, without having to impose a huge safety radius around the obstacle. The main point of analysis for a reactive local planner is on how the safety distance is maintained from the constrained environment throughout the experiments, which is shown in Figure \ref{fig:distance} for the NMPC and in Figure \ref{fig:pf_distance} for the APFs. These Figures represent the closest range measurement from the on-board 2D LiDAR throughout the flights, and as seen the minimum safety distance is maintained completely throughout the experiments for the NMPC, except a small \unit[0.03]{m} violation for the third experiment, while the APF has a slightly larger \unit[0.07]{m} violation also during the third experiment while moving through the small opening. The main difference here is the ability for the NMPC to generate control signals that navigate the UAV as to precisely satisfy the desired safety distance, due to the method of constraining the available position space. It should also be noted, that the safety distance for the NMPC is defined by the obstacle geometries, not the LiDAR measurements themselves, so there might not be a perfect overlap. For example, in the case of the first experiment of avoiding the cylinder, the minimum distance is \unit[0.43]{m}, while the constraint is satisfied precisely. 

\begin{figure}[ht]
\centering
  \includegraphics[width=0.8\linewidth]{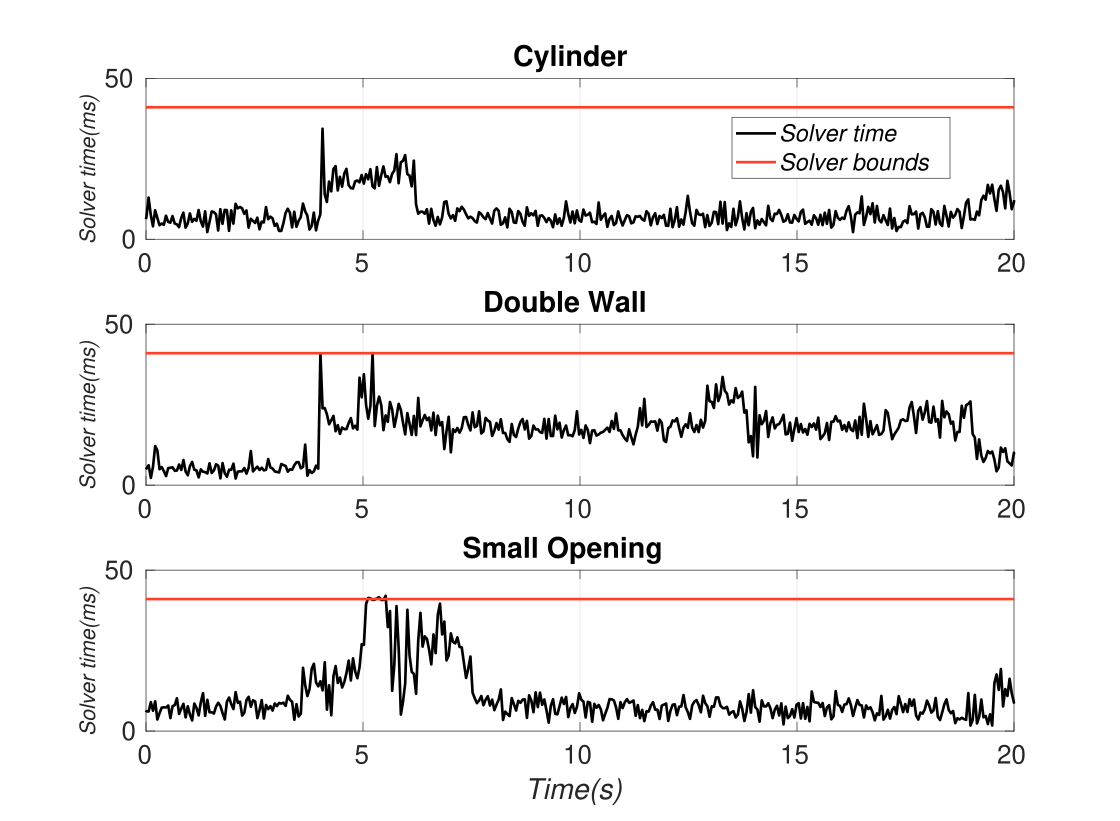}
  \caption{NMPC-module solver time for the three experiments.}
  \label{fig:solvertime}
\end{figure}

\begin{figure}[ht]
\centering
  \includegraphics[width=0.8\linewidth]{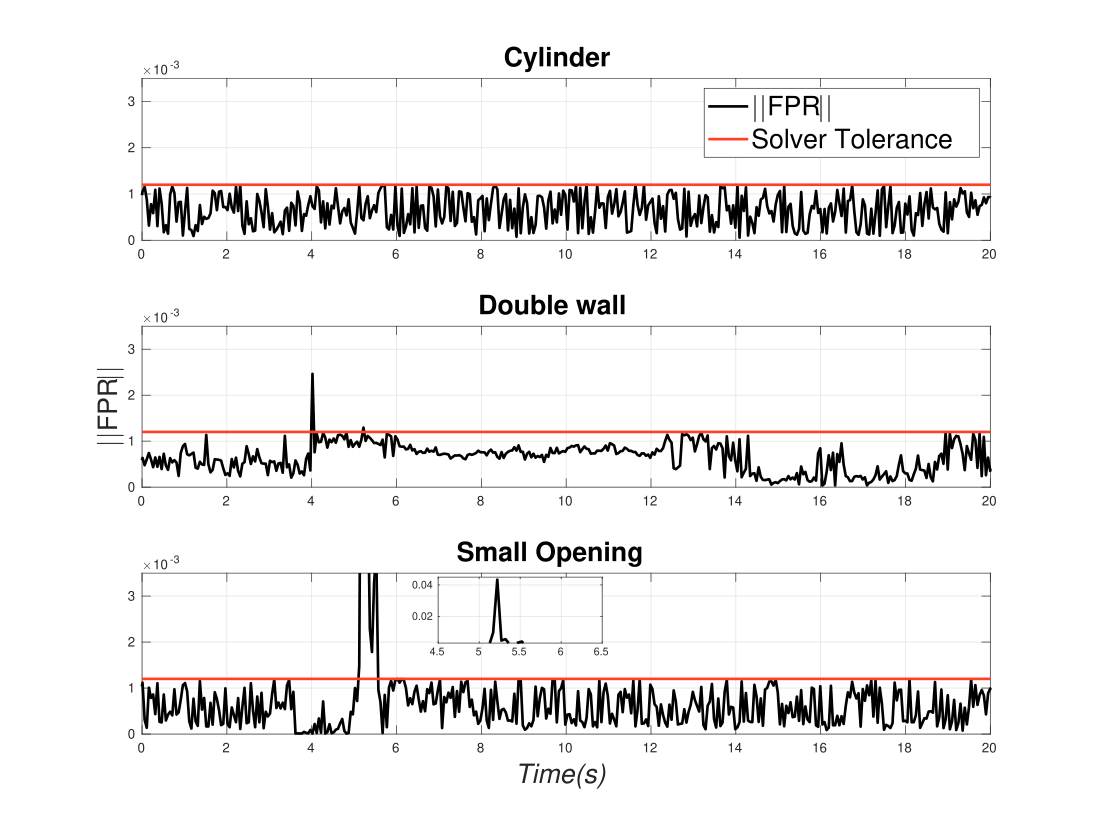}
  \caption{NMPC-module norm of the fixed-point residual for the last inner problem for the three experiments.}
  \label{fig:fpr}
\end{figure}

\begin{figure}[ht]
\centering
  \includegraphics[width=0.8\linewidth]{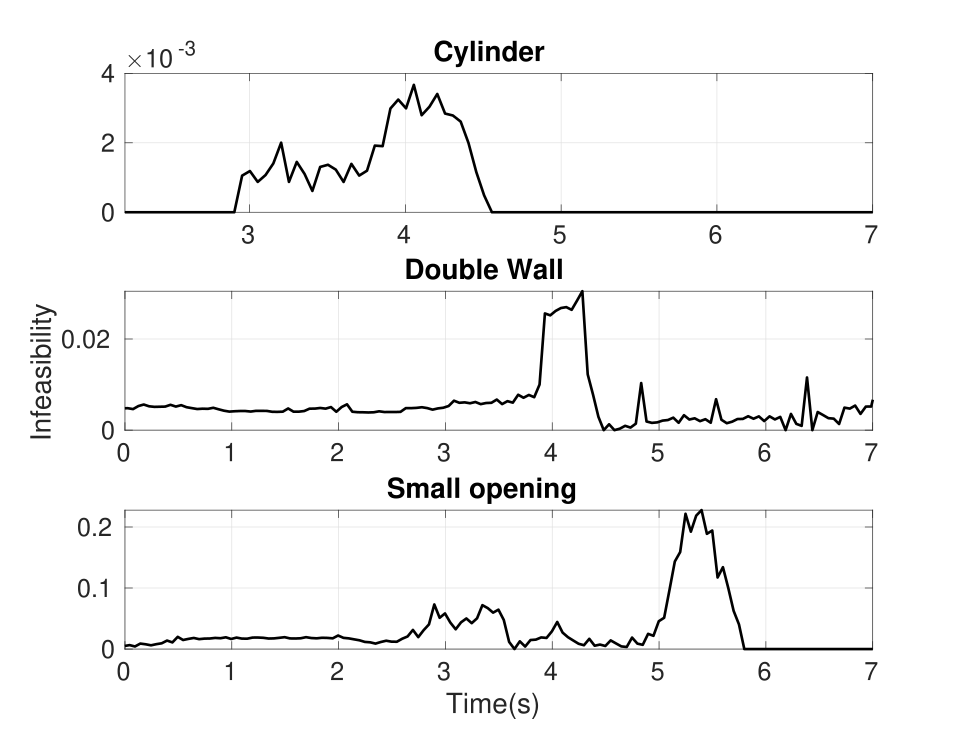}
  \caption{NMPC-module constraint infeasibility during critical time instant during avoidance maneuvers.}
  \label{fig:infeasibility}
\end{figure}

Figures \ref{fig:solvertime} and Figure \ref{fig:fpr} display the solver data for the three flights, namely the solver time of the NMPC-module and the norm of the fixed-point residual of the last inner problem, describing the the suboptimality of the solution (which here can be seen as a measurement of solver convergence, or as the quality of the solution). Figure \ref{fig:infeasibility} also shows the constraint infeasibility as the Euclidean norm of $G(z,\rho)$, during the critical part of the obstacle avoidance maneuver.
Measuring the average solver time during the avoidance maneuver (for example, between 4-7 seconds in the cylinder experiment) we get an average of \unit[17.8]{ms}, \unit[19.6]{ms} and \unit[22.3]{ms} respectively for the three scenarios, while we momentarily hit the bounds for the solver time in the latter two scenarios. The result of hitting this bound can be clearly observed in Figure \ref{fig:fpr} and \ref{fig:infeasibility} where the norm of the fixed-point residual peaks above the solver tolerance for these time instants, and we see a similar increase in the infeasibility. Generally, there is only one time instant where the optimizer is not close to converging, which is at \unit[5.2]{s} into the third experiment, a case presented in Figure \ref{fig:video_bad}. Despite this, since the trajectory is re-calculated at 20Hz intervals, a feasible solution is found at some moments later and the UAV still avoids the obstacle. 

\begin{figure}[ht]
\centering
  \includegraphics[width=0.8\linewidth]{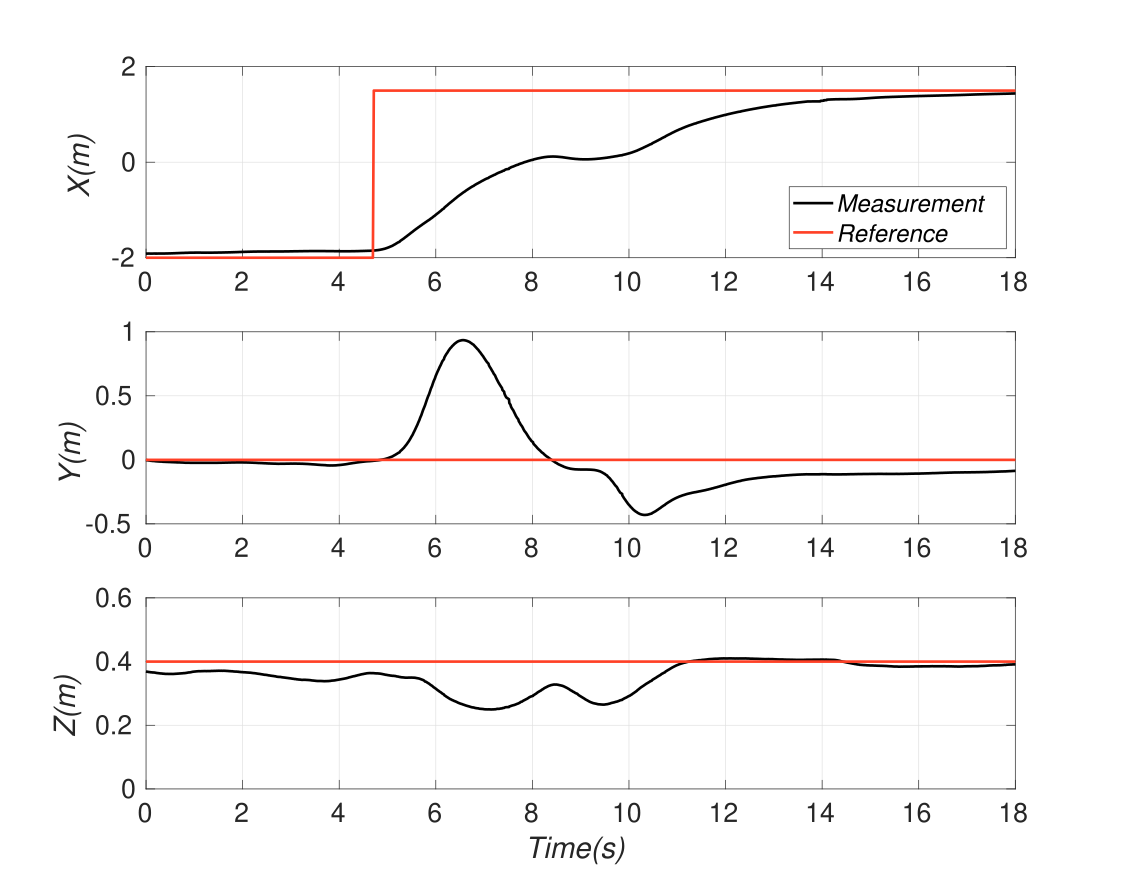}
  \caption{Reference tracking for the obstacle avoidance scenario with two walls.}
  \label{fig:2walls_pos}
\end{figure}

\begin{figure}[ht]
\centering
  \includegraphics[width=0.8\linewidth]{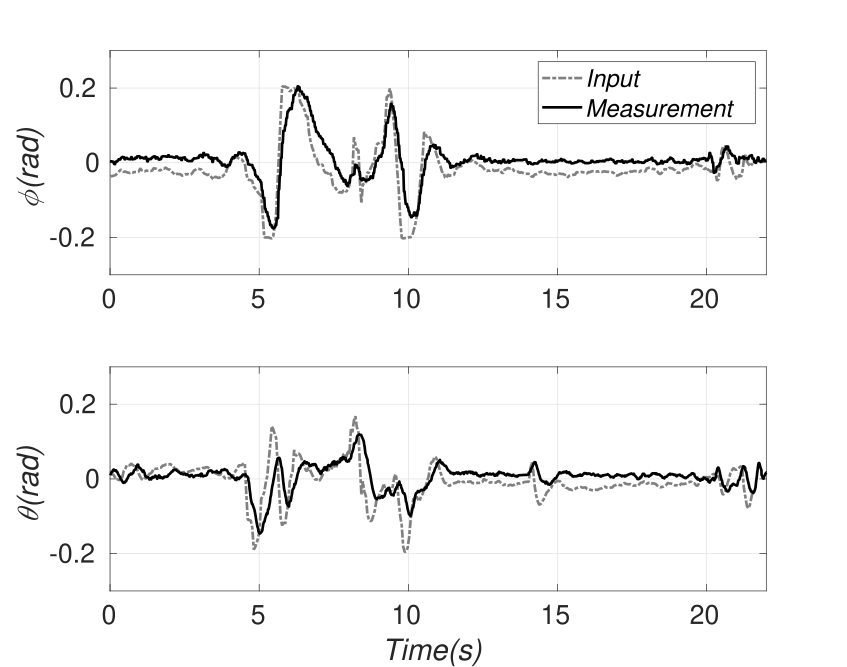}
  \caption{Computed control signals for the scenario with two walls, and corresponding angle states.}
  \label{fig:2walls_input}
\end{figure}

\begin{figure}[ht]
\centering
  \includegraphics[width=0.8\linewidth]{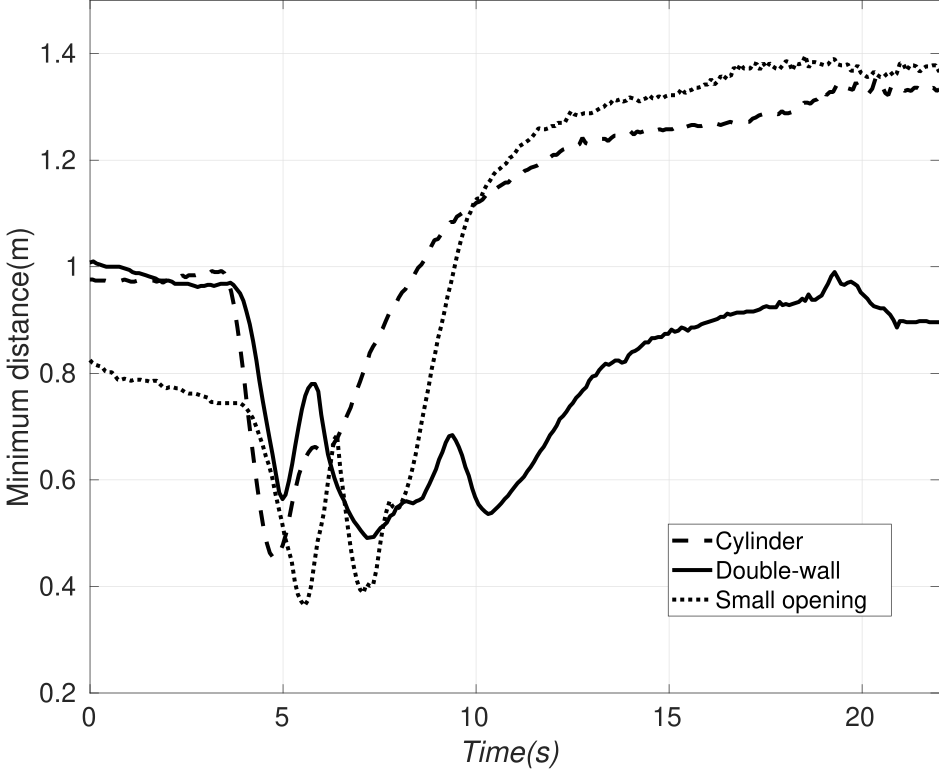}
  \caption{The minimum-distance from any obstacle by the closest LiDAR range measurement during the three experiment scenarios for the NMPC-based approach.}
  \label{fig:distance}
\end{figure}

\begin{figure}[ht]
\centering
  \includegraphics[width=0.9\linewidth]{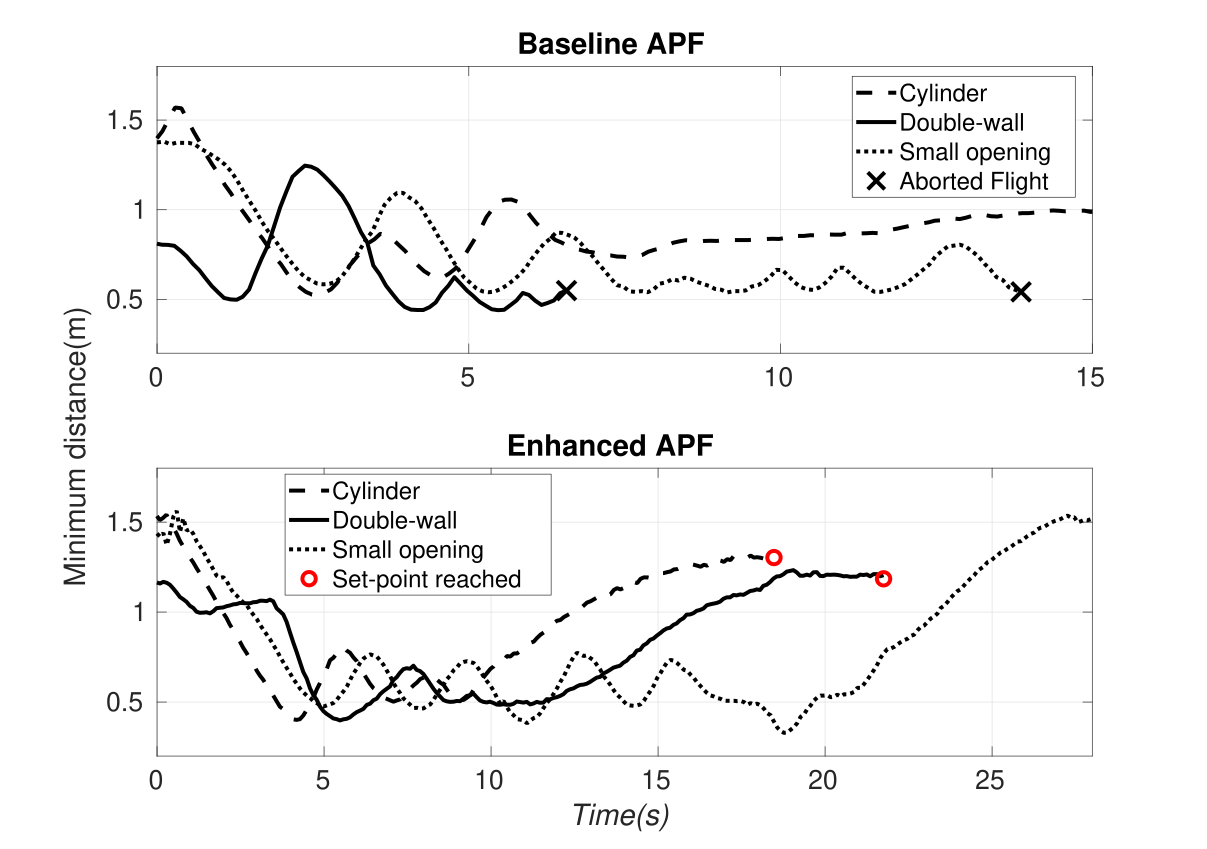}
  \caption{The minimum-distance from any obstacle by the closest LiDAR range measurement during the three experiment scenarios for the two APF approaches.}
  \label{fig:pf_distance}
\end{figure}

\begin{figure}[ht]
\centering
  \includegraphics[width=0.8\linewidth]{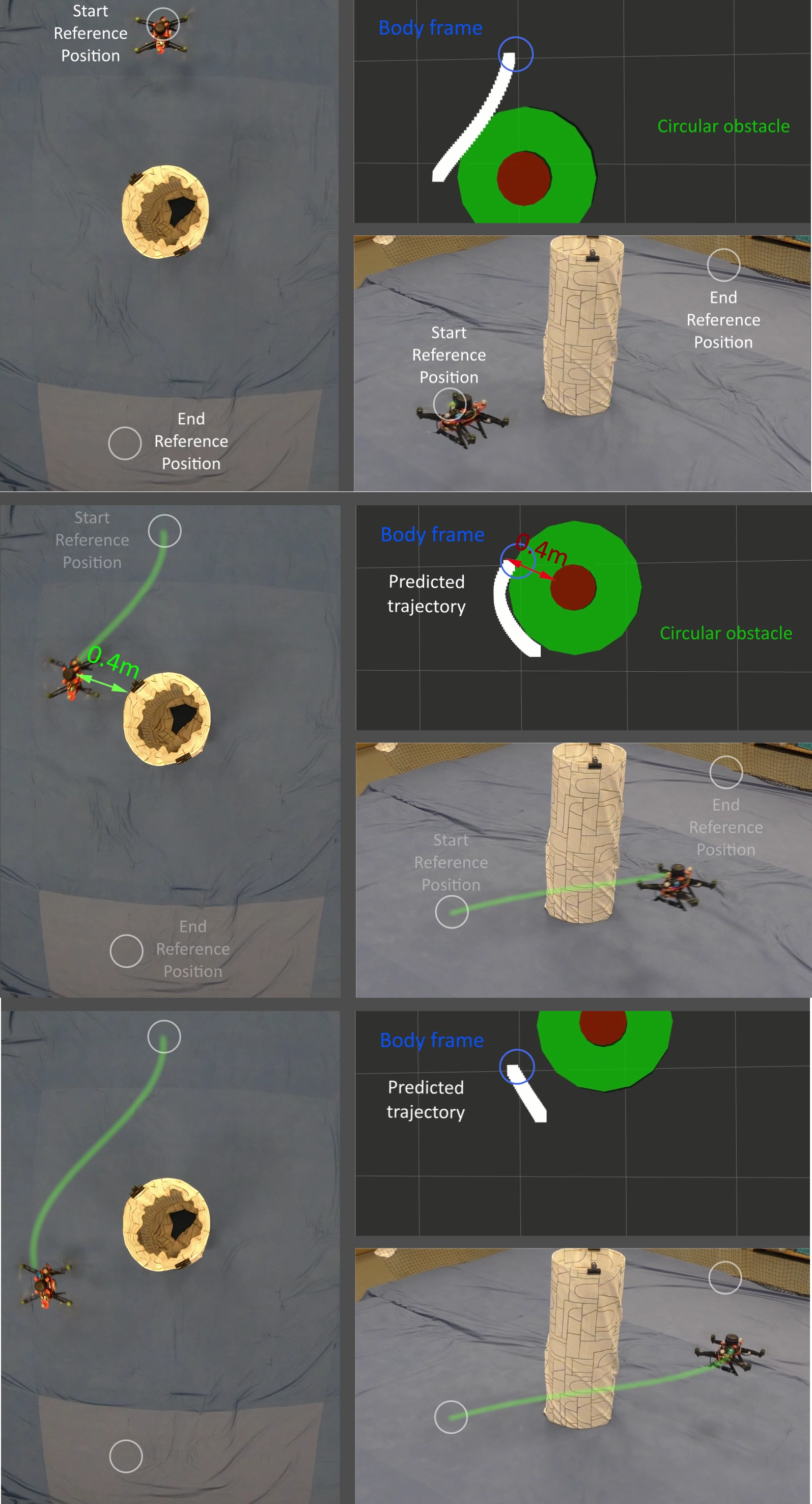}
  \caption{Snapshot images of the first experiment. The UAV is tasked to avoid a cylindrical obstacle while performing set-point tracking. The maximum size-radius of the UAV is 0.3m, as such we set the safety distance to 0.4m.}
  \label{fig:exp1_merged}
\end{figure}

Additionally, in Figure \ref{fig:2walls_pos} the set-point reference tracking from the double-wall experiment is displayed. From the obtained results it is obvious that the UAV cleanly deviates from the reference to perform the obstacle avoidance maneuver. Figure \ref{fig:2walls_input} shows the inputs and Euler angle states from the same experiment. These reflect the high input costs and the application of input and input rate constraints, where there is a relatively smooth behavior of the control inputs in $\phi_{\mathrm{ref}}, \theta_{\mathrm{ref}}$ despite the multi-obstacle constrained environment. 

\section{Limitations and Future Works}\label{sec:limitations}
\begin{figure}[ht!]
\centering
  \includegraphics[width=0.8\linewidth]{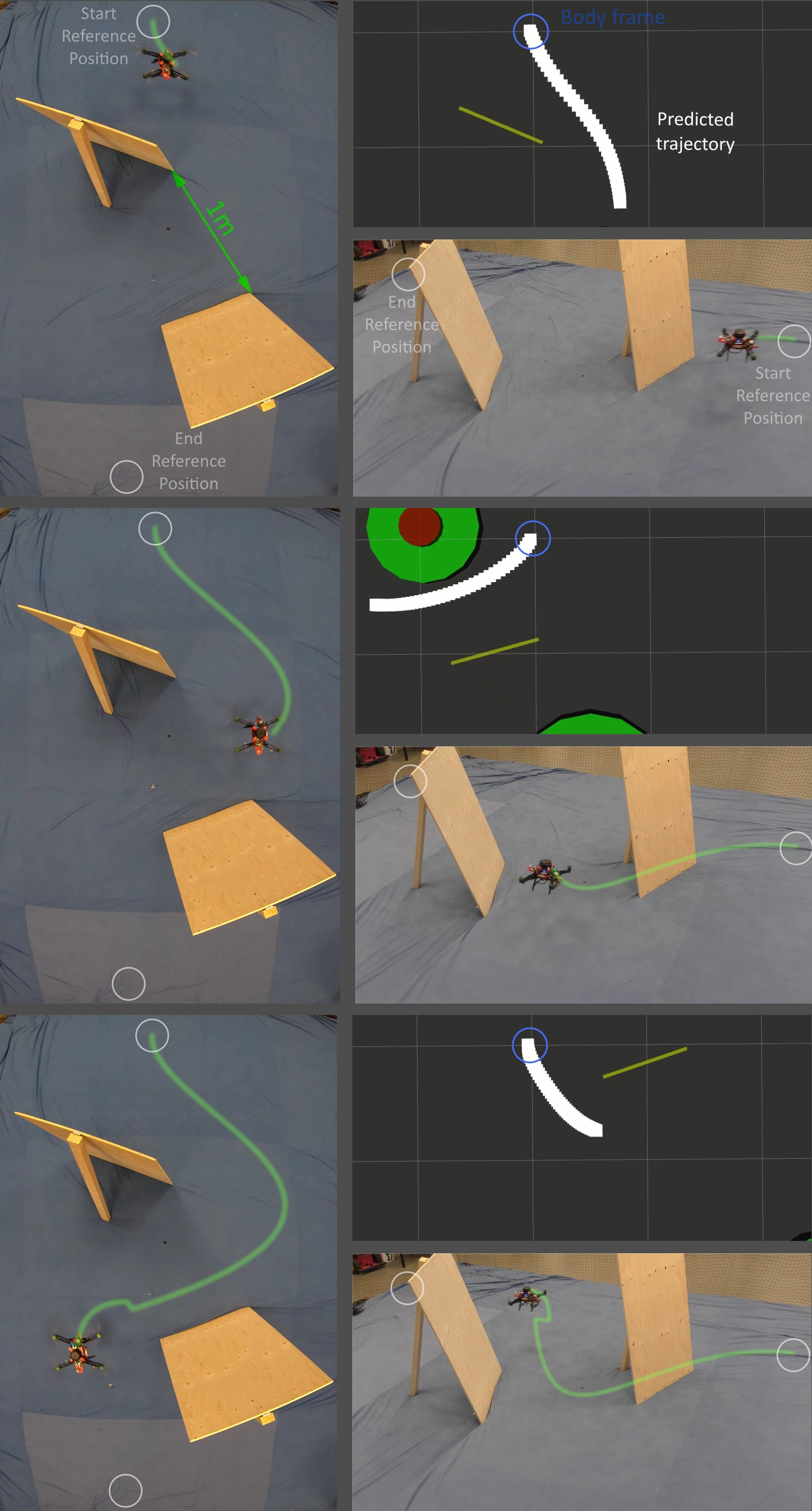}
  \caption{Snapshot images of the second experiment. The UAV is tasked to move through the multi-obstacle constrained environment and avoid any obstacles.}
  \label{fig:exp2_merged}
\end{figure}

\begin{figure}[ht]
\centering
  \includegraphics[width=0.8\linewidth]{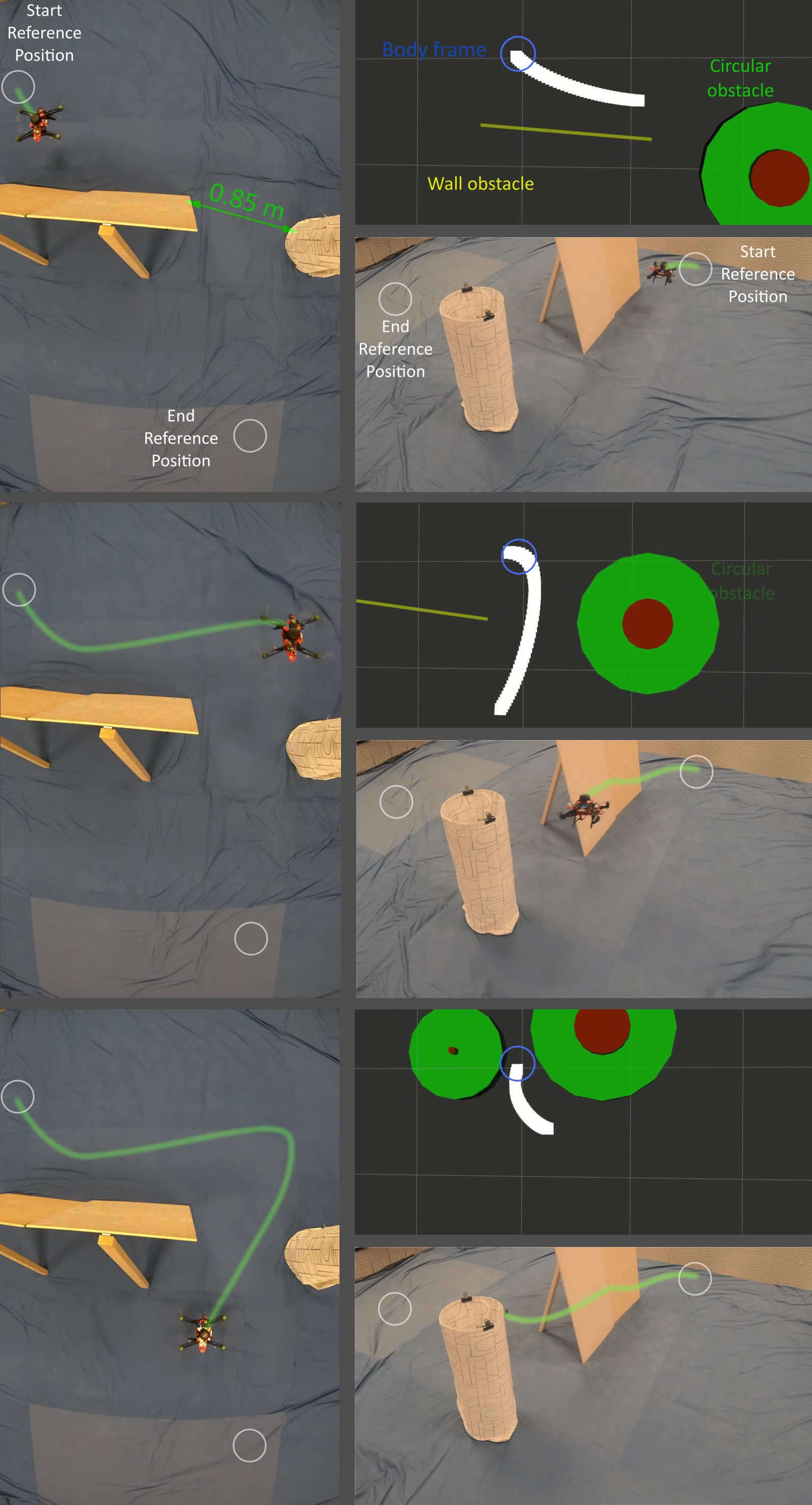}
  \caption{Snapshot images from the third experiment. The UAV is tasked to move to the set-point reference, where the free path is a small constrained opening of 0.85m between two obstacles. Using a safety radius of 0.4m the UAV can precisely pass through with an aggressive turn (within the dynamic constraints).}
  \label{fig:exp3_merged}
\end{figure}

\begin{figure}[ht]
\centering
  \includegraphics[width=0.7\linewidth]{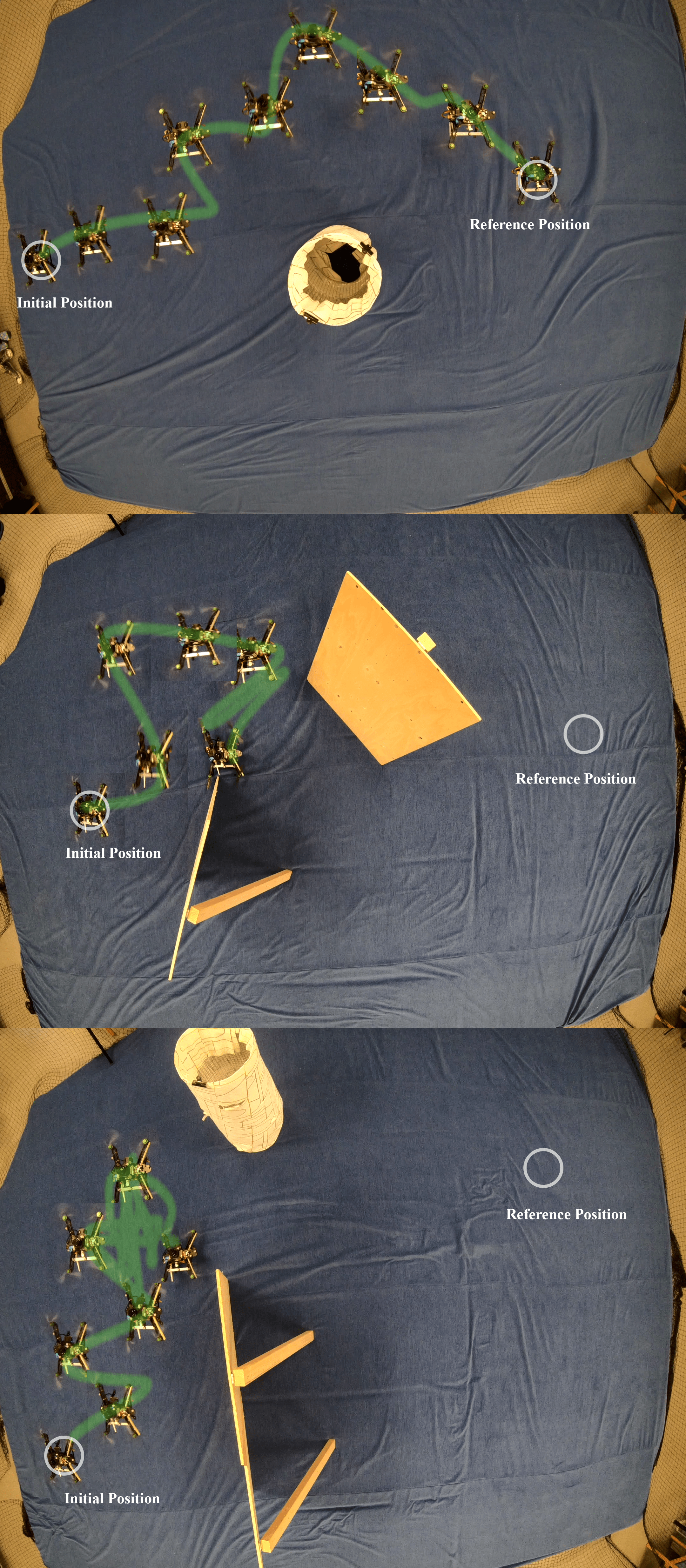}
  \caption{Paths produced by the baseline APF, for the three experiment scenarios. The scheme is incapable of completing the last two obstacle courses due to overly aggressive maneuvering.}
  \label{fig:pf_w_merged}
\end{figure}

\begin{figure}[ht]
\centering
  \includegraphics[width=0.7\linewidth]{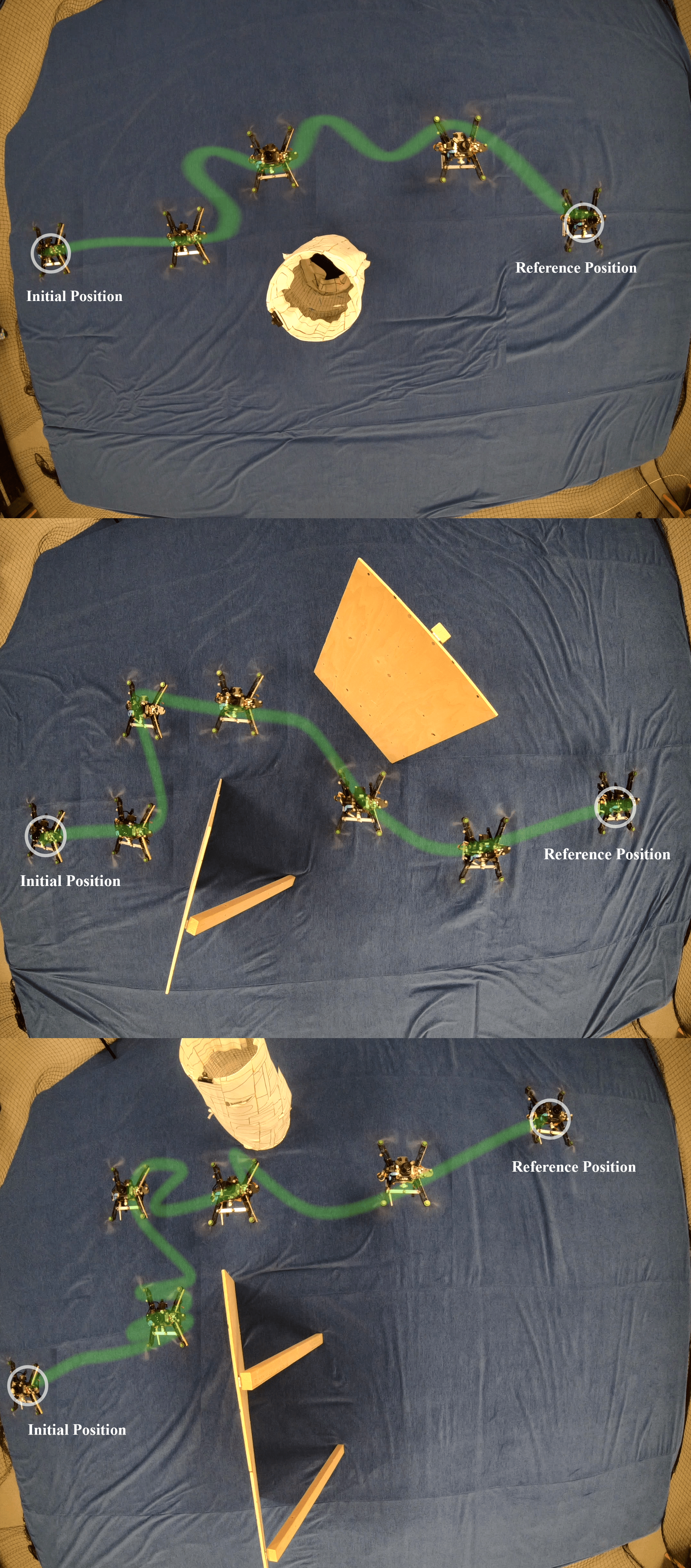}
  \caption{Paths produced by the enhanced APF, for the three experiment scenarios. The scheme completes all three scenarios, but struggles to move through the small opening.}
  \label{fig:pf_b_merged}
\end{figure}

\begin{figure}[ht]
\centering
  \includegraphics[width=0.8\linewidth]{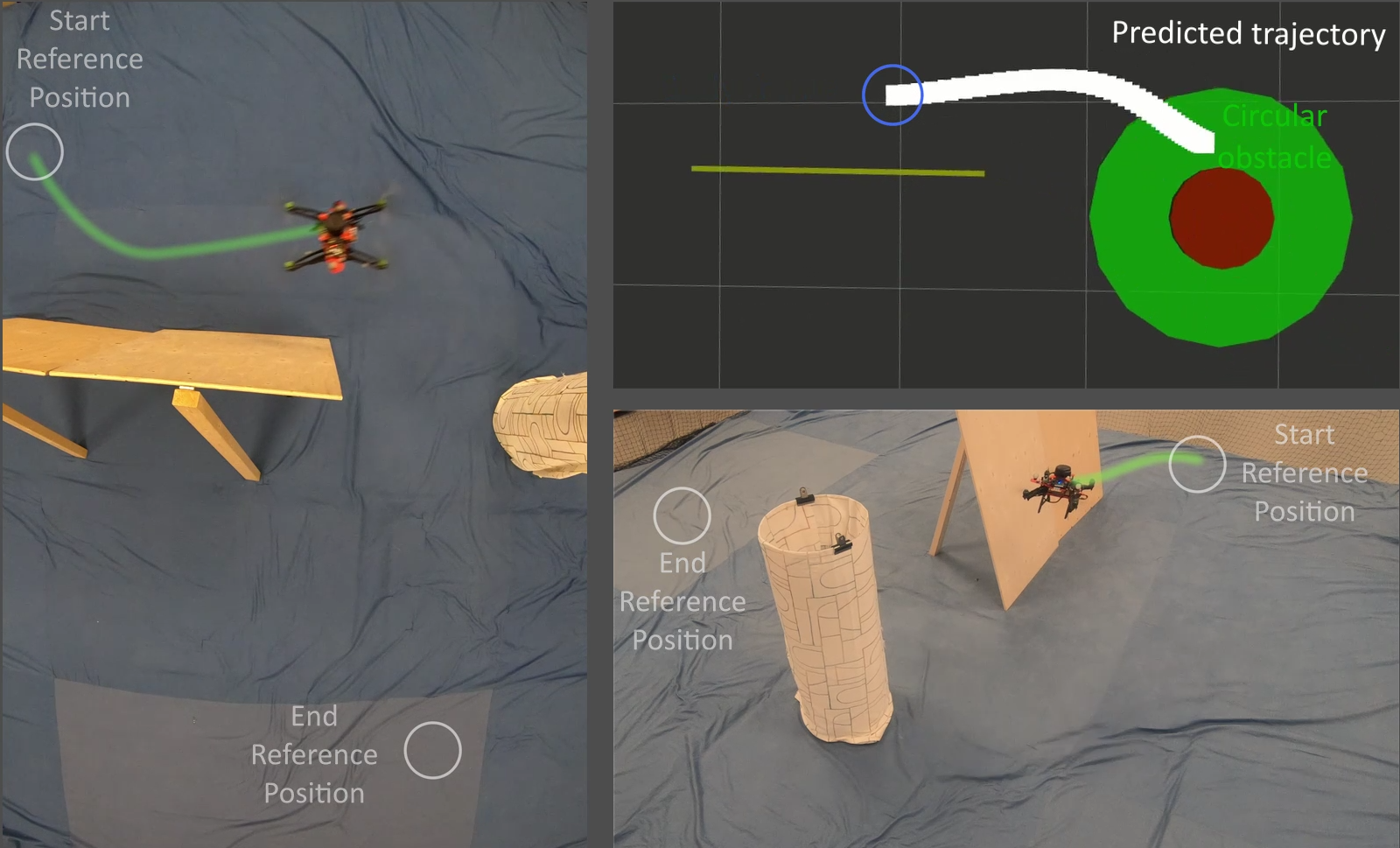}
  \caption{Image from the third experiment, showing the non-converged solution produced at 5.2s into the small-opening experiment in Figure \ref{fig:fpr}.}
  \label{fig:video_bad}
\end{figure}

Working with a solution to the obstacle avoidance and local path planning problem in terms of nonlinear MPC with perception-based parametric constraints, we have discovered some limitations to this framework that provide directions for future works. The biggest limitation is the reliance on geometric data of the environment that has to run in real-time and on-board. Small shifts in these geometric approximations can lead to large shifts in trajectory planning, where paths through the environment might close or open up due to these shifts, or the obstacle classification suddenly changes (either by improper approximation or from seeing the obstacle from a different angle) resulting in a different position-space being constrained by the NMPC, for example as middle figure of Figure \ref{fig:exp2_merged} where the combined stand plus wall are merged into a circle as the UAV passes the obstacle.
As such, proper filtering and tracking of the identified obstacles is a necessity for applying this method to avoid issues relating to obstacles suddenly changing positions, disappearing, or the filtering process replicating/duplicating certain obstacles due to fast enough changes in measurements when maneuvering so that the filter does not keep up (see results video). It should be noted that most methods (visual, distance to collision via convoluted neural networks~\cite{kouris2018learning}, occupancy-based etc.) have similar restrictions and requirements, with some exceptions like the artificial potential fields that can work directly with LiDAR data. 
Additionally, as this work is limited to using two obstacles types (circles and line segments), either developing algorithms that can classify more types of obstacles, or by using more general obstacles (by for example using segments of high-degree polynomials to define obstacles boundaries) is a clear limitation and direction of future work.

Also, as discussed in Section \ref{sec:results}, the solver might hit the bounds for computation time, which leads to non-converged (or not optimal) solutions of the NMPC problem being applied to the UAV system (which does not guarantee obstacle-free trajectories). Adding conditions to re-stabilize mid-flight and re-calculate, and perhaps relax some input constraints, when the solver does not converge would be a suitable addition to the method to guarantee that the trajectory $\bm{u_k}$ is obstacle-free before applying $u_{k \mid k}$ to the system. Further analysis and work on conditioning the constraints, and investigating other methods for constraint-based NMPC navigation, such as the Augmented Lagrangian Method~\cite{birgin2014practical} (which is implemented in OpEn, but would require a different approach to defining complex obstacles as in \eqref{eq:lineconstraint}), or the popular Barrier Functions~\cite{wills2004barrier} (which have previously been used in similar applications~\cite{chen2017obstacle}), could be utilized instead of the Penalty Method used in this article. 
Additionally, further developing the Penalty Method implementation to allow for different penalty update factors, initial penalties, and constraint tolerances for different constraints would supplement the very different constraints posed in Section \ref{sec:const} as, for example, the optimal parameters for the input rate constraints differ from the constraints that arise from obstacle avoidance.

The problems of solver time could also be prevented by running the method on more powerful on-board computers (which would also allow more computationally heavy mapping/environment awareness algorithms), such as the now-popular Intel NUC. Additionally, purely for future works, an important consideration is to extend the method to 3D obstacles, where obstacle data might come from depth-cameras or 3D LiDARs. Again, this would require reliable and fast algorithms for obstacle parameterization/approximation based on 3D camera or LiDAR data, and probably more powerful on-board computation. Another method for future investigation could be segmenting occupied space from an occupancy map representation\cite{hornung2013octomap} into obstacle geometries.

\section{Conclusions} \label{sec:conclusion}
In this article we have presented a novel obstacle avoidance method based on nonlinear model predictive control (NMPC), using the Optimization Engine (OpEn) used for an Unmanned Aerial Vehicle (UAV) in constrained environments. The method performs control, local path planning and reactive obstacle avoidance all integrated into the control layer and as such the optimized trajectories are within the dynamic constraints of the system, while being described by the dynamic system model. The efficacy of the control scheme has been demonstrated in multiple experimental scenarios, where the UAV avoids all obstacles, while completing the mission of set-point tracking. To the authors best knowledge this is the first time such a scheme has been applied on a real UAV, for a fully autonomous reactive navigation where the parametric NMPC constraints are derived online from the surrounding environment using on-board computation and sensors. 

The NMPC outperformed our Artificial Potential Field (APF) algorithms on the ability to execute the mission of set-point tracking as quickly as possible, as well as performing considerably more efficient avoidance maneuvers. While APFs are great at maintaining distance from obstacles as an additional reactive safety layer, the NMPCs ability to precisely satisfy safety distances and proactively initiate its avoidance, without the need to impose very large areas of influence of the avoidance, greatly enhances its ability to move in-between multiple close-by obstacles, or pass through tight openings, which becomes very significant in tightly constrained environments such as in subterranean or indoors urban environments.

As demonstrated, the NMPC successfully solves the optimization problem online with few exceptions during the challenging obstacle-avoidance scenarios, while maintaining input, input rate, and obstacle constraints. The proposed method is shown to handle multiple obstacle and it is also capable of passing through tight constrained spaces. Since the proposed complete architecture is totally novel, there are multiple directions of future works to improve the efficacy of the method, such as different methods for handling nonlinear non-convex constraints and extending the method to include 3D obstacles by using 3D LiDARs or depth-cameras for obstacle detection and parameterization. 

\bibliographystyle{IEEEtran.bst}
\bibliography{mybib}

\begin{thebibliography}{10}
\providecommand{\url}[1]{#1}
\csname url@samestyle\endcsname
\providecommand{\newblock}{\relax}
\providecommand{\bibinfo}[2]{#2}
\providecommand{\BIBentrySTDinterwordspacing}{\spaceskip=0pt\relax}
\providecommand{\BIBentryALTinterwordstretchfactor}{4}
\providecommand{\BIBentryALTinterwordspacing}{\spaceskip=\fontdimen2\font plus
\BIBentryALTinterwordstretchfactor\fontdimen3\font minus
  \fontdimen4\font\relax}
\providecommand{\BIBforeignlanguage}[2]{{%
\expandafter\ifx\csname l@#1\endcsname\relax
\typeout{** WARNING: IEEEtran.bst: No hyphenation pattern has been}%
\typeout{** loaded for the language `#1'. Using the pattern for}%
\typeout{** the default language instead.}%
\else
\language=\csname l@#1\endcsname
\fi
#2}}
\providecommand{\BIBdecl}{\relax}
\BIBdecl

\bibitem{mansouri2018cooperative}
S.~S. Mansouri, C.~Kanellakis, E.~Fresk, D.~Kominiak, and G.~Nikolakopoulos,
  ``Cooperative coverage path planning for visual inspection,'' \emph{Control
  Engineering Practice}, vol.~74, pp. 118--131, 2018.

\bibitem{mansouri2019visioncnn}
S.~S. Mansouri, P.~Karvelis, C.~Kanellakis, D.~Kominiak, and G.~Nikolakopoulos,
  ``Vision-based mav navigation in underground mine using convolutional neural
  network,'' in \emph{IECON 2019-45th Annual Conference of the IEEE Industrial
  Electronics Society}, vol.~1.\hskip 1em plus 0.5em minus 0.4em\relax IEEE,
  2019, pp. 750--755.

\bibitem{tomic2012toward}
T.~Tomic, K.~Schmid, P.~Lutz, A.~Domel, M.~Kassecker, E.~Mair, I.~L. Grixa,
  F.~Ruess, M.~Suppa, and D.~Burschka, ``Toward a fully autonomous uav:
  Research platform for indoor and outdoor urban search and rescue,''
  \emph{IEEE robotics \& automation magazine}, vol.~19, no.~3, pp. 46--56,
  2012.

\bibitem{zegre2018delivery}
J.~K. Z{\`e}gre-Hemsey, B.~Bogle, C.~J. Cunningham, K.~Snyder, and W.~Rosamond,
  ``Delivery of automated external defibrillators (aed) by drones: Implications
  for emergency cardiac care,'' \emph{Current cardiovascular risk reports},
  vol.~12, no.~11, p.~25, 2018.

\bibitem{darpa}
\BIBentryALTinterwordspacing
{DARPA Subterranean (SubT) Challenge }. Accessed: 2021-05-23. [Online].
  Available: \url{https://www.subtchallenge.com/}
\BIBentrySTDinterwordspacing

\bibitem{IEEEdarpa}
\BIBentryALTinterwordspacing
(2020) {Late Nights, Cool Hacks, and More Stories From the DARPA SubT Urban
  Circuit}. Accessed: 2021-05-23. [Online]. Available:
  \url{http://disq.us/t/3mld3js}
\BIBentrySTDinterwordspacing

\bibitem{costar}
\BIBentryALTinterwordspacing
{COSTAR}. Accessed: 2021-05-23. [Online]. Available:
  \url{https://costar.jpl.nasa.gov/}
\BIBentrySTDinterwordspacing

\bibitem{agha2021nebula}
A.~Agha, K.~Otsu, B.~Morrell, D.~D. Fan, R.~Thakker, A.~Santamaria-Navarro,
  S.-K. Kim, A.~Bouman, X.~Lei, J.~Edlund \emph{et~al.}, ``Nebula: Quest for
  robotic autonomy in challenging environments; team costar at the darpa
  subterranean challenge,'' \emph{arXiv preprint arXiv:2103.11470}, 2021.

\bibitem{kim2021plgrim}
S.-K. Kim, A.~Bouman, G.~Salhotra, D.~D. Fan, K.~Otsu, J.~Burdick, and A.-a.
  Agha-mohammadi, ``Plgrim: Hierarchical value learning for large-scale
  exploration in unknown environments,'' in \emph{Proceedings of the
  International Conference on Automated Planning and Scheduling}, vol.~31,
  2021, pp. 652--662.

\bibitem{palieri2020locus}
M.~Palieri, B.~Morrell, A.~Thakur, K.~Ebadi, J.~Nash, A.~Chatterjee,
  C.~Kanellakis, L.~Carlone, C.~Guaragnella, and A.-a. Agha-Mohammadi, ``Locus:
  A multi-sensor lidar-centric solution for high-precision odometry and 3d
  mapping in real-time,'' \emph{IEEE Robotics and Automation Letters}, vol.~6,
  no.~2, pp. 421--428, 2020.

\bibitem{lavalle2006planning}
S.~M. LaValle, \emph{Planning algorithms}.\hskip 1em plus 0.5em minus
  0.4em\relax Cambridge university press, 2006.

\bibitem{goerzen2010survey}
C.~Goerzen, Z.~Kong, and B.~Mettler, ``A survey of motion planning algorithms
  from the perspective of autonomous {UAV} guidance,'' \emph{Journal of
  Intelligent and Robotic Systems}, vol.~57, no. 1-4, p.~65, 2010.

\bibitem{kuffner2000rrt}
J.~J. Kuffner and S.~M. LaValle, ``Rrt-connect: An efficient approach to
  single-query path planning,'' in \emph{Proceedings 2000 ICRA. Millennium
  Conference. IEEE International Conference on Robotics and Automation.
  Symposia Proceedings (Cat. No. 00CH37065)}, vol.~2.\hskip 1em plus 0.5em
  minus 0.4em\relax IEEE, 2000, pp. 995--1001.

\bibitem{duchoe2014path}
F.~Ducho{\.E}, A.~Babineca, M.~Kajana, P.~Be{\.E}oa, M.~Floreka, T.~Ficoa, and
  L.~Juri{\v{s}}icaa, ``Path planning with modified a star algorithm for a
  mobile robot,'' \emph{Procedia Engineering}, vol.~96, pp. 59--69, 2014.

\bibitem{otte2016rrtx}
M.~Otte and E.~Frazzoli, ``Rrtx: Asymptotically optimal single-query
  sampling-based motion planning with quick replanning,'' \emph{The
  International Journal of Robotics Research}, vol.~35, no.~7, pp. 797--822,
  2016.

\bibitem{pharpatara20163}
P.~Pharpatara, B.~H{\'e}riss{\'e}, and Y.~Bestaoui, ``3-d trajectory planning
  of aerial vehicles using rrt,'' \emph{IEEE Transactions on Control Systems
  Technology}, vol.~25, no.~3, pp. 1116--1123, 2016.

\bibitem{rimon1992exact}
E.~Rimon and D.~E. Koditschek, ``Exact robot navigation using artificial
  potential functions,'' \emph{IEEE Transactions on Robotics and Automation},
  vol.~8, no.~5, pp. 501--518, 1992.

\bibitem{droeschel2016multilayered}
D.~Droeschel, M.~Nieuwenhuisen, M.~Beul, D.~Holz, J.~St{\"u}ckler, and
  S.~Behnke, ``Multilayered mapping and navigation for autonomous micro aerial
  vehicles,'' \emph{Journal of Field Robotics}, vol.~33, no.~4, pp. 451--475,
  2016.

\bibitem{kanellakis2018towards}
C.~Kanellakis, S.~S. Mansouri, G.~Georgoulas, and G.~Nikolakopoulos, ``Towards
  autonomous surveying of underground mine using mavs,'' in \emph{International
  Conference on Robotics in Alpe-Adria Danube Region}.\hskip 1em plus 0.5em
  minus 0.4em\relax Springer, Cham, 2018, pp. 173--180.

\bibitem{hrabar2011reactive}
S.~Hrabar, ``Reactive obstacle avoidance for rotorcraft uavs,'' in \emph{2011
  IEEE/RSJ International Conference on Intelligent Robots and Systems}.\hskip
  1em plus 0.5em minus 0.4em\relax IEEE, 2011, pp. 4967--4974.

\bibitem{oleynikova2016continuous}
H.~Oleynikova, M.~Burri, Z.~Taylor, J.~Nieto, R.~Siegwart, and E.~Galceran,
  ``Continuous-time trajectory optimization for online uav replanning,'' in
  \emph{2016 IEEE/RSJ International Conference on Intelligent Robots and
  Systems (IROS)}.\hskip 1em plus 0.5em minus 0.4em\relax IEEE, 2016, pp.
  5332--5339.

\bibitem{schaub2016reactive}
A.~Schaub, D.~Baumgartner, and D.~Burschka, ``Reactive obstacle avoidance for
  highly maneuverable vehicles based on a two-stage optical flow clustering,''
  \emph{IEEE Transactions on Intelligent Transportation Systems}, vol.~18,
  no.~8, pp. 2137--2152, 2016.

\bibitem{ruf2018real}
B.~Ruf, S.~Monka, M.~Kollmann, and M.~Grinberg, ``Real-time on-board obstacle
  avoidance for uavs based on embedded stereo vision,'' \emph{arXiv preprint
  arXiv:1807.06271}, 2018.

\bibitem{alexis2011switching}
K.~Alexis, G.~Nikolakopoulos, and A.~Tzes, ``Switching model predictive
  attitude control for a quadrotor helicopter subject to atmospheric
  disturbances,'' \emph{Control Engineering Practice}, vol.~19, no.~10, pp.
  1195--1207, 2011.

\bibitem{rosolia2016autonomous}
U.~Rosolia, S.~De~Bruyne, and A.~G. Alleyne, ``Autonomous vehicle control: A
  nonconvex approach for obstacle avoidance,'' \emph{IEEE Transactions on
  Control Systems Technology}, vol.~25, no.~2, pp. 469--484, 2016.

\bibitem{soloperto2019collision}
R.~Soloperto, J.~K{\"o}hler, F.~Allg{\"o}wer, and M.~A. M{\"u}ller, ``Collision
  avoidance for uncertain nonlinear systems with moving obstacles using robust
  model predictive control,'' in \emph{2019 18th European Control Conference
  (ECC)}.\hskip 1em plus 0.5em minus 0.4em\relax IEEE, 2019, pp. 811--817.

\bibitem{sathya2018embedded}
A.~Sathya, P.~Sopasakis, R.~Van~Parys, A.~Themelis, G.~Pipeleers, and
  P.~Patrinos, ``Embedded nonlinear model predictive control for obstacle
  avoidance using panoc,'' in \emph{2018 European Control Conference
  (ECC)}.\hskip 1em plus 0.5em minus 0.4em\relax IEEE, 2018, pp. 1523--1528.

\bibitem{stella2017simple}
L.~Stella, A.~Themelis, P.~Sopasakis, and P.~Patrinos, ``A simple and efficient
  algorithm for nonlinear model predictive control,'' in \emph{2017 IEEE 56th
  Annual Conference on Decision and Control (CDC)}.\hskip 1em plus 0.5em minus
  0.4em\relax IEEE, 2017, pp. 1939--1944.

\bibitem{small2019aerial}
E.~Small, P.~Sopasakis, E.~Fresk, P.~Patrinos, and G.~Nikolakopoulos, ``Aerial
  navigation in obstructed environments with embedded nonlinear model
  predictive control,'' in \emph{2019 18th European Control Conference
  (ECC)}.\hskip 1em plus 0.5em minus 0.4em\relax IEEE, 2019, pp. 3556--3563.

\bibitem{lindqvist2020nonlinear}
B.~Lindqvist, S.~S. Mansouri, and G.~Nikolakopoulos, ``Non-linear mpc based
  navigation for micro aerial vehicles in constrained environments,'' in
  \emph{2020 European Control Conference (ECC), May}, 2020.

\bibitem{Hermans:IFAC:2018}
B.~Hermans, P.~Patrinos, and G.~Pipeleers, ``A penalty method based approach
  for autonomous navigation using nonlinear model predictive control,''
  \emph{IFAC-PapersOnLine}, vol.~51, no.~20, pp. 234 -- 240, 2018.

\bibitem{hermans2021penalty}
B.~Hermans, G.~Pipeleers, and P.~P. Patrinos, ``A penalty method for nonlinear
  programs with set exclusion constraints,'' \emph{Automatica}, vol. 127, p.
  109500, 2021.

\bibitem{sharif2020subterranean}
S.~Sharif~Mansouri, C.~Kanellakis, E.~Fresk, B.~Lindqvist, D.~Kominiak,
  A.~Koval, P.~Sopasakis, and G.~Nikolakopoulos, ``Subterranean mav navigation
  based on nonlinear mpc with collision avoidance constraints,'' in
  \emph{International Federation of Automatic Control}, 2020.

\bibitem{open2019}
\BIBentryALTinterwordspacing
P.~Sopasakis, E.~Fresk, and P.~Patrinos, ``{Optimization Engine},'' 2019.
  [Online]. Available: \url{http://doc.optimization-engine.xyz/}
\BIBentrySTDinterwordspacing

\bibitem{sopasakis2020open}
------, ``Open: Code generation for embedded nonconvex optimization,''
  \emph{International Federation of Automatic Control}, 2020.

\bibitem{mansouri2020unified}
S.~S. Mansouri, C.~Kanellakis, B.~Lindqvist, F.~Pourkamali-Anaraki, A.-A.
  Agha-Mohammadi, J.~Burdick, and G.~Nikolakopoulos, ``A unified nmpc scheme
  for mavs navigation with 3d collision avoidance under position uncertainty,''
  \emph{IEEE Robotics and Automation Letters}, vol.~5, no.~4, pp. 5740--5747,
  2020.

\bibitem{kamel2017nonlinear}
M.~Kamel, J.~Alonso-Mora, R.~Siegwart, and J.~Nieto, ``Nonlinear model
  predictive control for multi-micro aerial vehicle robust collision
  avoidance,'' \emph{arXiv preprint arXiv:1703.01164}, 2017.

\bibitem{lindqvist2020dynamic}
B.~Lindqvist, S.~S. Mansouri, A.-a. Agha-mohammadi, and G.~Nikolakopoulos,
  ``Nonlinear mpc for collision avoidance and control of uavs with dynamic
  obstacles,'' \emph{IEEE Robotics and Automation Letters}, vol.~5, no.~4, pp.
  6001--6008, 2020.

\bibitem{jackson2016rosflight}
J.~{Jackson}, G.~{Ellingson}, and T.~{McLain}, ``{ROSflight: A lightweight,
  inexpensive MAV research and development tool},'' in \emph{2016 International
  Conference on Unmanned Aircraft Systems (ICUAS)}, June 2016, pp. 758--762.

\bibitem{warren1989global}
C.~W. Warren, ``Global path planning using artificial potential fields,'' in
  \emph{1989 IEEE International Conference on Robotics and Automation}.\hskip
  1em plus 0.5em minus 0.4em\relax IEEE Computer Society, 1989, pp. 316--317.

\bibitem{quigley2009ros}
M.~Quigley, K.~Conley, B.~Gerkey, J.~Faust, T.~Foote, J.~Leibs, R.~Wheeler, and
  A.~Y. Ng, ``{ROS}: an open-source robot operating system,'' in \emph{ICRA
  workshop on open source software}, vol.~3.\hskip 1em plus 0.5em minus
  0.4em\relax Kobe, Japan, 2009, p.~5.

\bibitem{kominiak2020mav}
D.~Kominiak, S.~S. Mansouri, C.~Kanellakis, and G.~Nikolakopoulos, ``Mav
  development towards navigation in unknown and dark mining tunnels,''
  \emph{arXiv preprint arXiv:2005.14433}, 2020.

\bibitem{mansouri2019vision}
S.~S. Mansouri, P.~Karvelis, C.~Kanellakis, D.~Kominiak, and G.~Nikolakopoulos,
  ``Vision-based mav navigation in underground mine using convolutional neural
  network,'' in \emph{IEEE Industrial Electronics Society}, 2019.

\bibitem{mansouri2020deploying}
S.~S. Mansouri, C.~Kanellakis, D.~Kominiak, and G.~Nikolakopoulos, ``Deploying
  mavs for autonomous navigation in dark underground mine environments,''
  \emph{Robotics and Autonomous Systems}, vol. 126, p. 103472, 2020.

\bibitem{przybyla2017detection}
M.~Przyby{\l}a, ``Detection and tracking of 2d geometric obstacles from lrf
  data,'' in \emph{2017 11th International Workshop on Robot Motion and Control
  (RoMoCo)}.\hskip 1em plus 0.5em minus 0.4em\relax IEEE, 2017, pp. 135--141.

\bibitem{kouris2018learning}
A.~Kouris and C.-S. Bouganis, ``Learning to fly by myself: A self-supervised
  cnn-based approach for autonomous navigation,'' in \emph{2018 IEEE/RSJ
  International Conference on Intelligent Robots and Systems (IROS)}.\hskip 1em
  plus 0.5em minus 0.4em\relax IEEE, 2018, pp. 1--9.

\bibitem{birgin2014practical}
E.~G. Birgin and J.~M. Mart\_nez, \emph{Practical augmented Lagrangian methods
  for constrained optimization}.\hskip 1em plus 0.5em minus 0.4em\relax SIAM,
  2014, vol.~10.

\bibitem{wills2004barrier}
A.~G. Wills and W.~P. Heath, ``Barrier function based model predictive
  control,'' \emph{Automatica}, vol.~40, no.~8, pp. 1415--1422, 2004.

\bibitem{chen2017obstacle}
Y.~Chen, H.~Peng, and J.~Grizzle, ``Obstacle avoidance for low-speed autonomous
  vehicles with barrier function,'' \emph{IEEE Transactions on Control Systems
  Technology}, vol.~26, no.~1, pp. 194--206, 2017.

\bibitem{hornung2013octomap}
A.~Hornung, K.~M. Wurm, M.~Bennewitz, C.~Stachniss, and W.~Burgard, ``Octomap:
  An efficient probabilistic 3d mapping framework based on octrees,''
  \emph{Autonomous robots}, vol.~34, no.~3, pp. 189--206, 2013.

\end{thebibliography}

\begin{IEEEbiography}
    [{\includegraphics[width=1in,height=1.25in,clip,keepaspectratio]{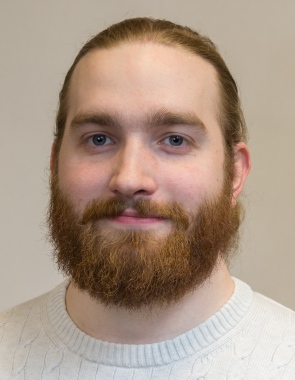}}]{Björn Lindqvist}
 is currently pursuing his PhD at the Robotics and AI Team at the Department of Computer Science, Electrical and Space Engineering, Luleå University of Technology, Sweden, working in the field of aerial robotics. He received his Master's Degree in Space Engineering with a specialisation Aerospace Engineering from Luleå University of Technology, Sweden, in 2019. Björn's research has so far been focused on collision avoidance and path planning for single and multi-agent Unmanned Aerial Vehicle systems, as well as field applications of such technologies. He has worked as part of the JPL-NASA led Team CoSTAR in the DARPA Sub-T Challenge on subterranean UAV exploration applications, specifically in the search-and-rescue context.
\end{IEEEbiography}

\begin{IEEEbiography}
    [{\includegraphics[width=1in,height=1.25in,clip,keepaspectratio]{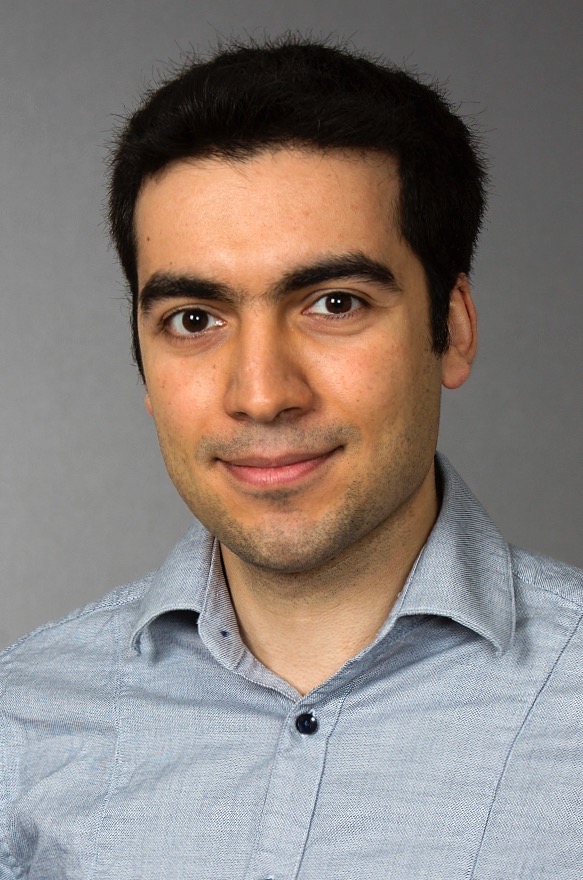}}]{Dr. Sina Sharif Mansouri}
is a post-doc at Robotics and AI team at Luleå University of Technology. He obtained his Phd degree within the Control Engineering Group, Department of Computer Science, Electrical and Space Engineering, Luleå University of Technology , Luleå, Sweden in 2020. He received his master of science and bachelor of science from Technical University of Dortmund, Germany in 2014 and University of Tehran, Iran in 2012 respectively. He currently works in the field of robotics, focusing on control, navigation and exploration with multiple agents. Sina has received the Vattenfall's award for best doctoral thesis 2021 and, his published scientific work includes more than 60 published International Journals and Conferences in the fields of his interest.
\end{IEEEbiography}

\begin{IEEEbiography}
    [{\includegraphics[width=1in,height=1.25in,clip,keepaspectratio]{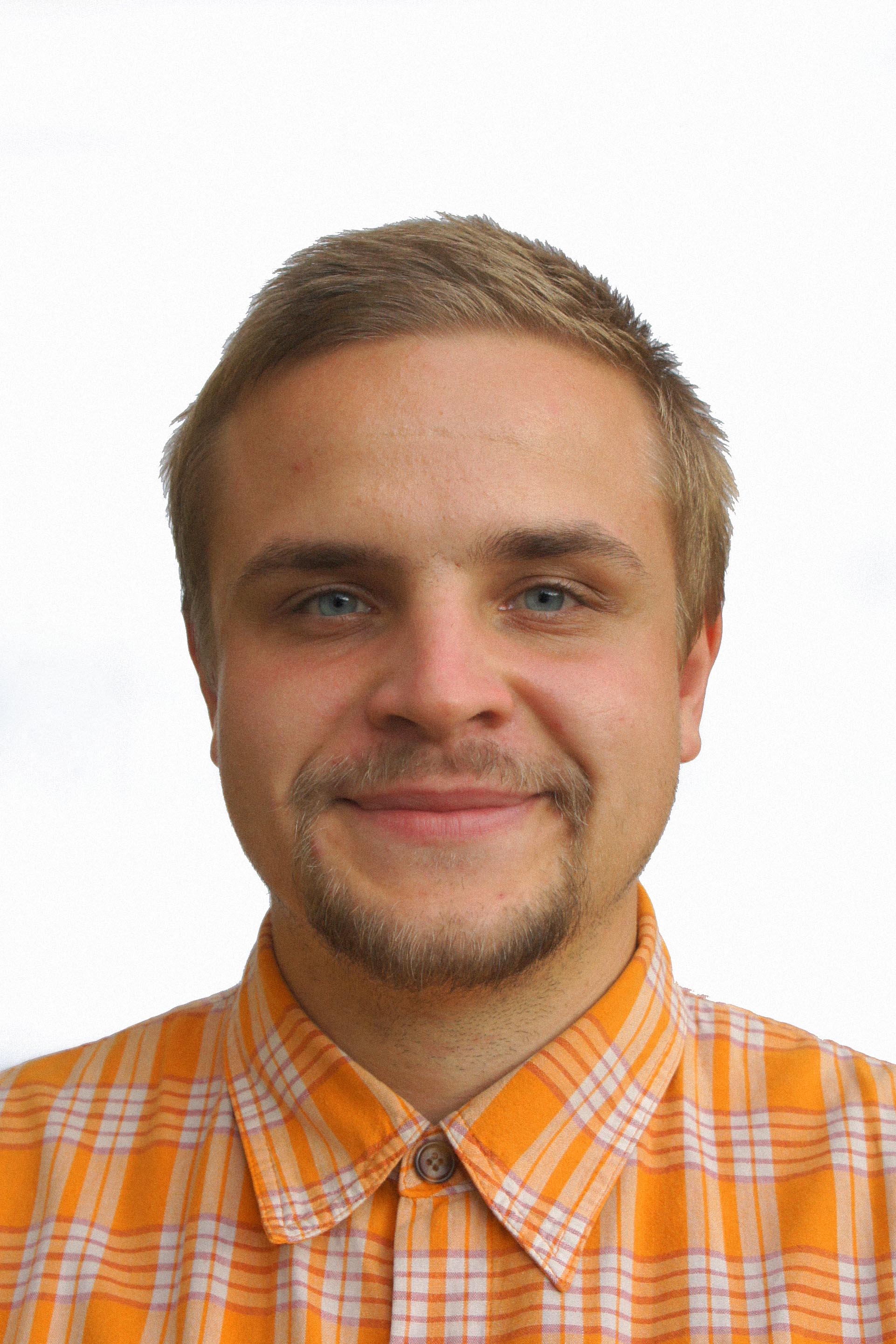}}]{Jakub Haluška}

received a Master's degree in Mechanical Engineering from the Technical University of Liberec, Czech Republic, in 2019. He is currently a Research Engineer within the Control Engineering Group, Department of Computer Science, Electrical and Space Engineering, Luleå University of Technology (LTU).  His focus is on the design and construction of robust and field-applicable robots as well as 3D printing technologies.
\end{IEEEbiography}

\begin{IEEEbiography}
    [{\includegraphics[width=1in,height=1.25in,clip,keepaspectratio]{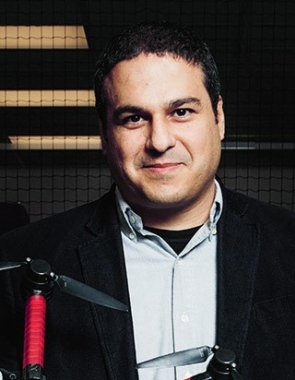}}]{Prof. George Nikolakopoulos}
works as the Chair Professor in Robotics and Artificial Intelligence, at the Department of Computer Science, Electrical and Space Engineering, Luleå University of Technology (LTU), Sweden. He was also affiliated with the NASA Jet Propulsion Laboratory for conducting collaborative research on Aerial Planetary Exploration and participated in the DARPA Grand challenge on Sub-T exploration with the CoSTAR team of JPL-NASA. He is a Director of euRobotics, and is a member of the Scientific Council of ARTEMIS and the IFAC TC on Robotics. He established the Digital Innovation Hub on Applied AI at LTU and also represents Sweden in the Technical Expert Group on Robotics and AI in the project MIRAI 2.0 promoting collaboration between Sweden and Japan. His main research interests are in the areas of: Field Robotics, Space Autonomy, UAVs, Automatic Control Applications, Networked Embedded Controlled Systems, Wireless Sensor and Actuator Networks, Cyber Physical Systems, and Adaptive Control.
\end{IEEEbiography}

\end{document}